\pdfoutput=1

\documentclass{article}

\usepackage{arxiv}

\usepackage[utf8]{inputenc} 
\usepackage[numbers,compress]{natbib}

\usepackage[T1]{fontenc}    
\usepackage{url}            
\usepackage{booktabs}       
\usepackage{amsfonts}       
\usepackage{nicefrac}       
\usepackage{microtype}      
\usepackage{xcolor}         
\usepackage{graphicx}      
\usepackage{amsmath}
\usepackage[most]{tcolorbox}
\usepackage{listings}
\usepackage{hyperref}       
\usepackage{wrapfig}
\usepackage{subcaption}
\usepackage{cleveref}       

\crefname{figure}{Fig.}{Figs.}
\Crefname{figure}{Figure}{Figures}

\crefname{table}{Tab.}{Tabs.}
\Crefname{table}{Table}{Tables}

\crefname{section}{Sec.}{Secs.}
\Crefname{section}{Section}{Sections}

\crefname{subsection}{Sec.}{Secs.}
\Crefname{subsection}{Section}{Sections}

\crefname{equation}{Eq.}{Eqs.}
\Crefname{equation}{Equation}{Equations}

\crefname{appendix}{App.}{Apps.}
\Crefname{appendix}{Appendix}{Appendices}

\newcommand{\abs}[1]{\left|#1\right|}

\definecolor{codepurple}{rgb}{0.5, 0.1, 0.6}
\definecolor{codegray}{rgb}{0.55, 0.55, 0.55}
\definecolor{boxbg}{rgb}{0.97, 0.97, 0.98}
\newcommand{\detached}[1]{\textcolor{codepurple}{#1}}

\usepackage[acronym]{glossaries}
\glsdisablehyper

\newacronym{actmap}{ActMap}{Activation Map}
\newacronym{ai}{AI}{Artificial Intelligence}
\newacronym{dl}{DL}{Deep Learning}
\newacronym{dnn}{DNN}{Deep Neural Network}
\newacronym{diffae}{DiffAE}{Diffusion Auto-Encoder}
\newacronym{lrp}{LRP}{Layer-wise Relevance Propagation}
\newacronym{lsb}{LSB}{least significant bit}
\newacronym{xai}{XAI}{eXplainable Artificial Intelligence}
\newacronym{crp}{CRP}{Concept Relevance Propagation}
\newacronym{amax}{ActMax}{Activation Maximization}
\newacronym{rmax}{RelMax}{Relevance Maximization}
\newacronym{auc}{AUC}{Area Under Curve}
\newacronym{llm}{LLM}{Large Language Models}
\newacronym{aoc}{AOC}{Area Over Curve}
\newacronym{conv}{Conv}{convolutional}
\newacronym{svm}{SVM}{Support Vector Machine}
\newacronym{roi}{ROI}{Region of Interest}
\newacronym{lcrp}{L-CRP}{CRP for Localization Models}
\newacronym{rrr}{RRR}{Right for the Right Reason}
\newacronym{cdep}{CDEP}{Contextual Decomposition Explanation Penalization}
\newacronym{clarc}{ClArC}{Class Artifact Compensation}
\newacronym{aclarc}{\mbox{A-ClArC}}{Augmentive ClArC}
\newacronym{pclarc}{\mbox{P-ClArC}}{Projective ClArC}
\newacronym{rrclarc}{RR-ClArC}{Right Reason ClArC}
\newacronym{ml}{ML}{Machine Learning}
\newacronym{cse}{CSE}{complete skin examination}
\newacronym{cav}{CAV}{Concept Activation Vector}
\newacronym{tcav}{TCAV}{Testing with CAV}
\newacronym{spray}{SpRAy}{Spectral Relevance Analysis}
\newacronym{iterrev}{IterRev}{Iteratively Revealing and Revising Spurious Model Behavior}
\newacronym{r2r}{R2R}{Reveal to Revise}
\newacronym{xil}{XIL}{eXplanatory Interactive Learning}
\newacronym{sem}{SEM}{Standard Error of the Mean}
\newacronym{se}{SE}{Standard Error}
\newacronym{sae}{SAE}{Sparse Autoencoder}
\newacronym{srg}{SRG}{Symmetric Relevance Gain}
\newacronym{vit}{ViT}{Vision Transformer}
\newacronym{cnn}{{CNN}}{Convolutional Neural Network}
\newacronym{ixg}{{I$\times$G}}{Input$\times$Grad}
\newacronym{ig}{{IG}}{Integrated Gradients}
\newacronym{legrad}{{LeGrad}}{Layerwise Explainability Gradient}
\newacronym{dave}{DAVE}{Distribution-aware Attribution via ViT Gradient Decomposition}
\newacronym{attnlrp}{AttnLRP}{Attention-aware Layer-wise Relevance Propagation}
\newacronym{cplrp}{CP-LRP}{Conservative Propagation LRP}
\newacronym{clrp}{Chefer-LRP}{C-LRP}
\newacronym{fullgrad}{{FullGrad}}{Full-Gradient Saliency Maps for Convolutional Nets}
\newacronym{abslrp}{{absLRP}}{Relative Absolute Magnitude Layer-Wise Relevance Propagation}
\newacronym{ours}{ResLRP}{Residual-aware Layer-wise Relevance Propagation}
\newacronym{vlm}{VLM}{Vision Language Model}
\newacronym{imagenet}{IN1K}{ImageNet 1K}
\newacronym{imagenets}{IN-S}{ImageNet-S}

\usepackage{fontawesome5}

\hypersetup{
    colorlinks=false, 
    pdfborder={0 0 0} 
}

\title{ResLRP: The Role of Residual Cancellation in Attribution Instability in Vision Transformers}

\renewcommand{\shorttitle}{Attribution Instability in Vision Transformers}

\author{
    Jim Berend$^{1}$\thanks{The authors contributed equally.}
    \quad
    Reduan Achtibat$^{1*}$
    \quad
    Daniel Sch\"affer$^{1}$
    \quad
    Alexander Binder$^{3,5,6}$
    \\
    \textbf{Wojciech Samek$^{1,2,4}$}
    \quad
    \textbf{Sebastian Lapuschkin$^{1}$}
    \quad
    \textbf{Maximilian Dreyer$^{1}$}
    \\[4pt]
    $^1$Fraunhofer Heinrich Hertz Institute
    \quad
    $^2$Technische Universit\"at Berlin
    \quad
    $^3$DSC ScaDS.AI, Leipzig University\\
    $^4$BIFOLD -- Berlin Institute for the Foundations of Learning and Data\\
    $^5$ICT Cluster, Singapore Institute of Technology, Singapore\\
    $^6$Institute for Cancer Genetics and Informatics (ICGI), Oslo, Norway\\
    \texttt{maximilian.dreyer@hhi.fraunhofer.de}
    \\[8pt]
    \faGithub\;\href{https://github.com/jim-berend/ResLRP}{\texttt{jim-berend/ResLRP}}
}

\date{}

\hypersetup{
    pdftitle={ResLRP: The Role of Residual Cancellation in Attribution Instability in Vision Transformers},
    pdfauthor={Jim Berend, Reduan Achtibat, Daniel Schaeffer, Alexander Binder, Wojciech Samek, Sebastian Lapuschkin, Maximilian Dreyer},
    pdfsubject={cs.CV, cs.LG},
    pdfkeywords={explainable AI, layer-wise relevance propagation, vision transformers, attribution, residual connections},
}

\begin{document}

\maketitle

\begin{abstract}
  \glspl{vit} are central to most modern vision models, yet obtaining input attributions that are fine-grained, faithful, and stable remains challenging.
  \gls{lrp} has been adapted to transformer attention, but in \glspl{vit} it often produces noisy, unfaithful explanations.
  We show that the missing ingredient is the treatment of residual connections: cancellation effects in residual pathways lead to attribution explosion.
  Moreover, we find that these cancellations are substantially stronger in \glspl{vit} than in language transformers.
  To address this issue, we introduce \gls{ours}, a simple extension of \gls{lrp} whose
  propagation rules explicitly account for cancellations in residual branches, are exactly
  conservative, and provably bound relevance explosion.
  Causal channel-wise interventions confirm that residual cancellation, not a generic
  regularization effect, drives the instability.
  \gls{ours} substantially improves attribution quality across faithfulness and localization, evaluated on \gls{vit} architectures spanning supervised, self-supervised, contrastive, hierarchical, and multimodal families, as well as on the ground-truth-controlled FunnyBirds benchmark.
  The largest gains arise in modern \glspl{vlm}, with
  $+27{-}29\%$ localization and up to $3.4\times$ faithfulness scores. Beyond benchmarks, \gls{ours}
  localizes \gls{sae} features in input space, and our residual amplification measure serves as an
  architecture-level diagnostic predicting where attribution degrades.

\end{abstract}

\section{Introduction}

\begin{figure}[t]
  \centering
  \includegraphics[width=1.0\linewidth]{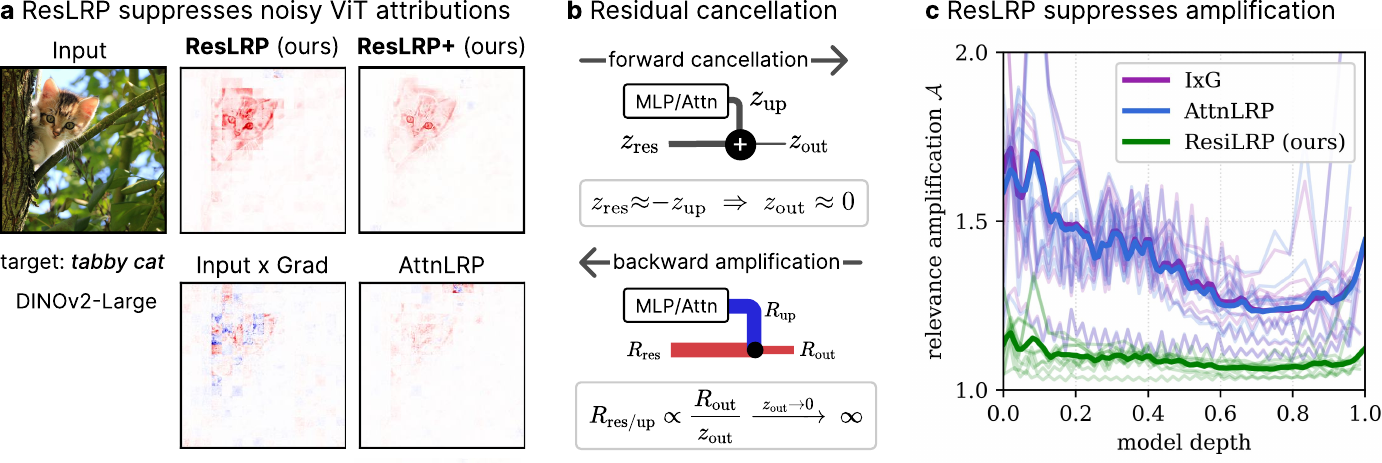}
  \caption{\gls{ours} mitigates attribution noise from residual cancellations.
    \textbf{a)} Cleaner heatmaps and more precise localization than other methods, smoothing further
    when combined with DAVE~\cite{dave}, see \cref{sec:resilrp-plus}.
    \textbf{b)} A cancellation occurs when an attention or MLP update opposes the residual stream,
    leaving a small post-addition activation. Standard rules divide by it, amplifying opposing
    relevance terms.
    \textbf{c)} \gls{ours} reduces amplification per residual site, most strongly in early layers
    (\cref{sec:motivation}).} \label{fig:main_intro}
\end{figure}

Transformer architectures dominate machine learning, powering state-of-the-art models in natural
language processing, computer vision, and multimodal
learning~\cite{vaswani2017attention,radford2021learning}. As they are increasingly deployed in
scientific and high-stakes applications, understanding which input features drive their predictions
becomes an important challenge~\cite{chefer2021transformer}. Input attributions, visualized as
``heatmaps'', are among the most widely used tools for this~\cite{bach2015pixel}, supporting
mechanistic analyses, concept-based interpretation, bias detection, and model
debugging~\cite{achtibat2023attribution,saeed2023explainable,weber2023beyond}. Gradient- and
backpropagation-based methods are the most practical of these, as they scale efficiently to large
models. For modern vision transformers, however, attributions that are both efficient and faithful
remain difficult to obtain. Heatmaps often contain checkerboard artifacts, diffuse relevance, or
fragmented high-frequency patterns that obscure the model's decision process, as illustrated in
\cref{fig:main_intro}a with Input$\times$Grad~\cite{shrikumar2017learning} and
\gls{attnlrp}~\cite{achtibat2024attnlrp}.

\acrfull{lrp} provides a principled framework for propagating prediction relevance through deep
networks~\cite{montavon2017explaining}, and recent work has extended \gls{lrp} rules to transformer
attention~\cite{achtibat2024attnlrp,rezaei2024mambalrp,bakish2025revisiting}. Despite these
advances, \gls{lrp}-based explanations for \glspl{vit} remain unstable and visually noisy. In this
work, we argue that a key source of this instability is not the treatment of attention alone, but
the treatment of \emph{residual connections}.

Residual connections are essential for optimization and representation learning in deep
networks~\cite{he2016deep}. In \glspl{vit}, however, we find that attention or MLP updates
frequently oppose the incoming residual stream, partially canceling information and producing small
combined activations. Such cancellations may be useful in the forward pass, where they remove,
refine, or contrast features, but when large opposing contributions cancel, standard propagation
rules assign relevance through a small effective denominator and amplify positive--negative
relevance pairs. We refer to this effect as \emph{relevance explosion}, which yields inflated,
brittle, and fragmented attribution maps (\cref{fig:main_intro}). It is substantially stronger in
\glspl{vit} than in language transformers and most pronounced in early layers, where
representations remain sensitive to local image structure~\cite{raghu2021vision}, so propagation
rules that behave stably in language models transfer poorly to \glspl{vit}.

To address this, we introduce \emph{\gls{ours}}, a residual-aware extension of \gls{lrp} that
explicitly accounts for cancellations between the residual stream and attention or MLP updates,
bounding and suppressing relevance assigned to contradictory pathways and thereby yielding cleaner,
more faithful maps (\cref{fig:main_intro}a). Beyond standard benchmarks, we demonstrate its value
in two settings of growing importance, localizing sparse autoencoder features back to input space
and explaining large vision--language models. The latter is where faithful attribution matters
most, as \glspl{vlm} are rapidly becoming the dominant deployed vision architecture, and exhibit
the strongest residual cancellation we measure. As predicted by our analysis, they also show the
largest gains from \gls{ours}.

In summary, our contributions are as follows: (1) We identify residual-path cancellation as a key
source of noisy and unstable \gls{lrp}-based attributions in \glspl{vit}, substantially stronger
than in language transformers and concentrated in early layers, and establish its causal role
through targeted channel-wise interventions and component ablations. (2) We introduce
\emph{\gls{ours}}, a simple residual-aware extension of \gls{lrp} that accounts for cancellation in
skip connections, is exactly conservative, and provably bounds cancellation-induced relevance
amplification. (3) We show that \gls{ours} improves faithfulness and localization across
supervised, self-supervised, contrastive, hierarchical, and multimodal architectures, on the
ground-truth-controlled FunnyBirds benchmark, on three modern \glspl{vlm}, and in \gls{sae} feature
localization. (4) We show that our residual amplification measure acts as an architecture-level
diagnostic, revealing that register tokens stabilize the residual stream itself and predicting
where attribution degrades, most notably in \glspl{vlm}.

\section{Related Work}

\textbf{Input attributions for vision.} Gradient-based methods such as saliency~\cite{Simonyan14a},
\gls{ixg}~\cite{shrikumar2017learning}, and \gls{ig}~\cite{sundararajan17} are efficient but prone
to gradient shattering~\cite{balduzzi17}, yielding noisy heatmaps, while perturbation surrogates
such as LIME~\cite{ribeiro2016should} and SHAP~\cite{lundberg2017unified} are more robust but
require hundreds of forward passes per explanation.

\textbf{Explainability for transformers.} Initial transformer methods focused on attention, such as
Attention Rollout~\cite{abnar20}, gradient-weighted attention~\cite{chefer2021transformer}, or
\gls{legrad}~\cite{bousselham25}, while the \gls{fullgrad} family~\cite{srinivas19, mehri24,
  mehri25} aggregates intermediate gradients and bias terms with targeted pruning to restore gradient
balance. Attention-based variants can be coarse or non-class-specific, and gradient-aggregation
methods are often not easily adapted to architecture variations such as missing \texttt{cls}-token
or hierarchical processing.

\textbf{LRP and the missing residual link.} Unlike raw gradients, \gls{lrp} modifies the backward
gradient layer-by-layer for better stability via dedicated rules (see \cite{montavon2019layer} for
an overview). Recent works successfully adapted \gls{lrp} to transformer models:
\gls{cplrp}~\cite{ali2022xai} and \gls{attnlrp}~\cite{achtibat2024attnlrp, bakish2025revisiting}
tackle attention non-linearities, \gls{abslrp}~\cite{vukadin24} corrects relative attribution
magnitudes, and \gls{lrp} has enabled circuit discovery in NLP~\cite{kahardipraja2026atlas,jafari2025relp, hatefi2025attribution}. However,
these advances primarily target attention and NLP models. By overlooking \emph{residual pathways},
prior works miss that standard \gls{lrp} attributions may degrade because of destructive
interference present in \gls{vit} residual streams. {Related in spirit, register
    tokens~\cite{Darcet2023VisionTN} mitigate attention artifacts in the forward pass at training time,
    whereas we correct cancellation-induced instability in the backward pass of pretrained models. As
    we show in \cref{sec:registers}, both perspectives are connected through the residual stream.}

\section{Residual Cancellations and Relevance Explosion in Vision Transformers}
\label{sec:motivation}

Transformer blocks repeatedly update the residual stream via additions $z_{\mathrm{out}} =
  z_{\mathrm{in}} + z_{\mathrm{up}}$, where $z_{\mathrm{in}}$ is the stream entering a sub-layer and
$z_{\mathrm{up}}$ the update produced by an attention or MLP block (\cref{fig:main_intro}b). While
residual connections preserve gradient flow and enable deeper networks, they induce a failure mode
for relevance propagation when the two branches oppose each other. We say a \emph{residual
  cancellation} occurs when $z_{\mathrm{in}} \approx -z_{\mathrm{up}}$, so that $z_{\mathrm{out}}$ is
small in magnitude despite both branches having individually large magnitudes: the forward activation hides
substantial internal computation.

\paragraph{Relevance propagation through a residual addition.}
\Gls{lrp} explains a prediction by redistributing relevance from the target output logit back to
the inputs layer by layer, assigning each input a share proportional to its forward contribution
and thereby conserving total relevance~\cite{bach2015pixel,montavon2019layer}. Specialized rules
such as the $\varepsilon$- and $\gamma$-rules improve stability in deep networks, and we refer to
prior work for a comprehensive treatment~\cite{montavon2019layer}. At a residual addition, the
classical $\varepsilon$-stabilized rule gives
\begin{equation}
  R_{\mathrm{in}}
  = \frac{z_{\mathrm{in}}}{z_{\mathrm{out}} + \operatorname{sign}(z_{\mathrm{out}})\,\varepsilon}
  R_{\mathrm{out}},
  \qquad
  R_{\mathrm{up}}
  = \frac{z_{\mathrm{up}}}{z_{\mathrm{out}} + \operatorname{sign}(z_{\mathrm{out}})\,\varepsilon}
  R_{\mathrm{out}}.
\end{equation}
The denominator is the residual output \emph{after} a potential cancellation, so a small
$|z_{\mathrm{out}}|$ inflates both redistribution factors and turns a small amount of output
relevance into large opposing values on the two branches. We call this \emph{relevance
  amplification}, or \emph{relevance explosion} in the high-amplification regime. The propagation is
thus ill-conditioned at cancellations, reacting highly sensitive to perturbations of the forward activations
and to the choice of $\varepsilon$. As a result the attributions can be dominated by local
cancellation structure, consuming attribution mass that would otherwise expose weaker but stable
relevance patterns persisting across layers~\cite{bohle2022b}.

\paragraph{Quantifying cancellation and amplification.}
To quantify forward cancellation at a residual update and the relevance mass it creates in the
backward pass, we define, over latent entries $i$,
\begin{equation}
  \mathcal{C}
  = \frac{\sum_i \left( |z_{\mathrm{in},i}| + |z_{\mathrm{up},i}| \right)}{\sum_j |z_{\mathrm{out},j}|},
  \qquad
  \mathcal{A}
  = \frac{\sum_i \left( |R_{\mathrm{in},i}| + |R_{\mathrm{up},i}| \right)}{\sum_j |R_{\mathrm{out},j}|}.
\end{equation}
Both compare the summed pre-addition magnitude of the two residual branches with the magnitude
remaining after the addition, in the forward pass for $\mathcal{C}$ and in the backward pass for
$\mathcal{A}$. Large $\mathcal{C}$ indicates that signal magnitude present before the addition is
suppressed by destructive interference, with $\mathcal{C} = 1$ meaning no cancellation, while
$\mathcal{A} \gg 1$ indicates that a small amount of output relevance gives rise to large opposing
relevance contributions in the preceding branches.

\begin{figure}[t]
  \centering
  \includegraphics[width=1.0\linewidth]{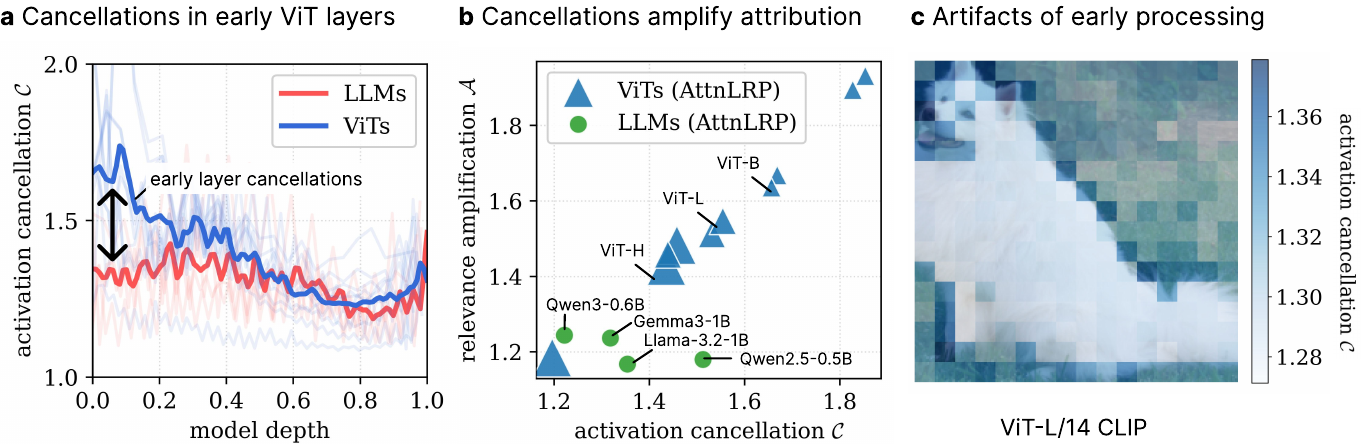}
  \caption{Residual cancellations are a primary source of relevance amplification in \glspl{vit}.
    \textbf{a)} Cancellations peak in early \gls{vit} layers and are weaker in language transformers.
    \textbf{b)} Stronger cancellation implies larger amplification under standard \gls{attnlrp},
    disproportionately affecting \glspl{vit}. Marker size denotes parameter count.
    \textbf{c)} In a ViT-L/14LIP model, block-2 cancellations activate on edges and the dog’s face
    (\cref{app:cancellation}).}
  \label{fig:main_motivation}
\end{figure}

Comparing residual sites across \glspl{vit} and \glspl{llm} (details in \cref{app:cancellation}),
we find pronounced cancellation in the early layers of \glspl{vit}, reaching $\mathcal{C} \approx
  1.7$ against $\mathcal{C} \approx 1.4$ for \glspl{llm} across layers (\cref{fig:main_motivation}a).
Across \glspl{vit}, $\mathcal{C}$ and $\mathcal{A}$ exhibit a Pearson correlation of $0.99$
(\cref{fig:main_motivation}b), so residual cancellations coincide with an increase in total
absolute relevance during backpropagation.

Since both quantities are structurally coupled through the residual propagation rule, this
correlation motivates our hypothesis but does not establish causality on its own. We therefore
verify the causal role of residual cancellation through targeted interventions in
\cref{sec:causal}. We further note that cancellation is necessary but not sufficient for relevance
explosion. Under the standard rule, $R_{\mathrm{in}} = \frac{z_{\mathrm{in}}}{z_{\mathrm{out}}}
  R_{\mathrm{out}}$, amplification depends on both the forward activations and the downstream
relevance $R_{\mathrm{out}}$. If cancelled directions are unused downstream, with $R_{\mathrm{out}}
  = 0$, no explosion occurs at $R_{\mathrm{in}}$. This dissociation is visible in \cref{fig:main_motivation}b, where
Qwen2.5 exhibits strong cancellation but little amplification.

These observations motivate us to introduce \gls{ours} and a dedicated propagation rule at residual
additions in the next section.

\section{Residual-Aware Layer-Wise Relevance Propagation}
Motivated by the high relevance amplification observed at residual sites in \glspl{vit}, we next
modify relevance propagation through residual additions. A desirable rule should preserve relevance
conservation while avoiding amplification along branches whose contributions cancel in the forward
pass. We therefore introduce \emph{\acrfull{ours}}, a residual-aware extension of the previous
\gls{attnlrp}~\cite{achtibat2024attnlrp} method, by additionally suppressing and bounding relevance
amplification induced by forward cancellations.

A residual addition takes two inputs $z_\text{in}, z_\text{up} \in \mathbb{R}$ and produces
$z_\text{out} := z_\text{in} + z_\text{up}$. Under the standard \gls{lrp} rule with
$\varepsilon=0$, relevance is propagated as $R_{\text{in}} = \frac{z_\text{in}}{z_\text{out}}
  R_{\text{out}}$, with $R_{\text{up}}$ defined analogously. As shown above, this becomes unstable
when the two branches cancel, redistributing a small amount of output relevance into large
contradictory positive--negative pairs that dominate subsequent propagation and overshadow more
stable attribution signals. In contrast, when both branches contribute with the same sign as the
output, the corresponding propagation factors remain well behaved.

To mitigate this effect, we adapt the \gls{lrp}-$\gamma$ rule ~\cite{montavon2019layer} to residual
additions. While this rule is usually applied to parametrized linear layers, we use it here to
favor branches whose contribution has the same sign as the residual output. Let $\delta_P$ denote
the indicator of statement $P$. Unless stated otherwise, $R_{\text{up}}$ is obtained from
$R_{\text{in}}$ by swapping the subscripts $\textit{in}$ and $\textit{up}$:
\begin{align}
  \label{eq:resilrp}
  \text{(\gls{lrp}-}\gamma\text{)}
  \qquad
  R_\text{in}
   & =
  \frac{
    \left(1+\gamma\,
    \delta_{\operatorname{sign}(z_\text{in})
      =
      \operatorname{sign}(z_\text{out})}\right) z_\text{in}
  }{
    \left(1+\gamma\,
    \delta_{\operatorname{sign}(z_\text{in})
      =
      \operatorname{sign}(z_\text{out})}\right) z_\text{in}
    +
    \left(1+\gamma\,
    \delta_{\operatorname{sign}(z_\text{up})
      =
      \operatorname{sign}(z_\text{out})}\right) z_\text{up}
  }
  R_\text{out}.
\end{align}
By strengthening sign-consistent contributions, this rule suppresses relevance
assigned to contradictory branches and, as shown next, bounds
cancellation-induced relevance amplification.
\gls{ours} requires only a single backward hook at each residual addition. A
complete PyTorch implementation is given in \cref{app:implementation}.
This rule is highly robust to the choice of $\gamma$; based on the hyperparameter sweep in \cref{sec:gamma}, we choose $\gamma=1$ and use it throughout this work.

\subsection{Theoretical Guarantees}
\label{methods:boundness}

\textbf{Proposition (Boundedness).} Let $a,b \in \mathbb{R}$ with $c := a+b \neq 0$ and $\gamma >
  0$, denote by $\delta_P \in \{0,1\}$ the indicator of statement $P$, and let
\begin{equation}
  \label{eq:phi-def}
  \Phi(a,b) := \frac{\left(1+\gamma\,\delta_{\operatorname{sign}(a)=\operatorname{sign}(c)}\right) a}
  {\left(1+\gamma\,\delta_{\operatorname{sign}(a)=\operatorname{sign}(c)}\right) a
    + \left(1+\gamma\,\delta_{\operatorname{sign}(b)=\operatorname{sign}(c)}\right) b}
\end{equation}
denote the factor scaling $R_\text{out}$ in \cref{eq:resilrp}. Then it holds
$-\tfrac{1}{\gamma} \le \Phi(a,b) \le 1+\tfrac{1}{\gamma}$. A proof is given in \cref{app:proof}.

\textbf{Corollary (Conservation and bounded amplification).} The symmetry $\Phi(a,b) + \Phi(b,a) =
  1$ follows directly from \cref{eq:phi-def} and, with \cref{eq:resilrp}, implies exact conservation
$R_{\mathrm{in}} + R_{\mathrm{up}} = R_{\mathrm{out}}$ at every residual addition for all $\gamma >
  0$, so \gls{ours} inherits the conservation properties of \gls{attnlrp}. Together with the
proposition this yields the tight per-merge bound
\begin{equation}
  \label{eq:tight-bound}
  \abs{R_{\mathrm{in}}} + \abs{R_{\mathrm{up}}} \le \left(1 + \tfrac{2}{\gamma}\right) \abs{R_{\mathrm{out}}},
\end{equation}
capping the measured amplification at $\mathcal{A} \le 1 + 2/\gamma$ per site
($\mathcal{A} \le 3$ at the default $\gamma = 1$), and at $(1+2/\gamma)^{2L}$ over the $2L$ residual
additions of an $L$-block transformer, monotonically decreasing in $\gamma$. The standard rule
admits no finite bound at even a single cancellation site.

The parameter $\gamma$ interpolates continuously between standard \gls{lrp}, and thus
\gls{attnlrp}, for $\gamma \to 0$ and a non-amplifying, sign-consistent redistribution for $\gamma
  \to \infty$, with a monotonically tightening bound (\cref{sec:gamma}). Both guarantees are exact
but local to the residual additions, and structural rather than faithfulness guarantees. Full-model
attribution quality is established empirically in
\cref{sec:quantitative,sec:causal,sec:funnybirds}.

\section{Experiments and Results}
\Cref{sec:quantitative} benchmarks faithfulness and localization against a broad set of attribution
methods and validates against controlled ground-truth part importance on
FunnyBirds~\cite{hesse2023funnybirds}. \Cref{sec:causal} then establishes causally that the gains
stem from correcting residual cancellation, \cref{sec:sae-attribution} demonstrates compatibility
with component-level attribution targets on \gls{sae} features, and \cref{sec:vlm-main} evaluates
three modern \glspl{vlm}. The appendix analyses sensitivity to the single hyperparameter $\gamma$
(\cref{sec:gamma}) and shows that residual amplification acts as an architecture-level diagnostic
(\cref{sec:registers}).

\subsection{Quantitative Evaluation of Attribution Quality}
\label{sec:quantitative} We compare \gls{ours} against a broad set of attribution methods across
seven checkpoints spanning supervised, self-supervised, contrastive, and hierarchical pre-training
regimes on images of the ImageNet dataset~\cite{Deng2009imagenet}. Experimental details are given
in \cref{sec:setup,app:baselines}, additional evaluation results in the appendix.

\subsubsection{Faithfulness: Symmetric Relevance Gain}

A faithful attribution should provide the input regions the model relies on for its prediction. We
measure this with \gls{srg}~\cite{blucher2024decoupling}, an occlusion-based benchmark that
requires no human annotations and directly probes the model's own decision process. Patches are
ranked by their attribution score and progressively occluded in two orders: most-important-first
(MIF) and least-important-first (LIF), with occluded regions replaced by the per-channel ImageNet
mean. A faithful attribution causes a steep logit drop under MIF and a slow drop under LIF;
\gls{srg} is the normalised area between these two curves: positive for a faithful method, zero for
a random baseline, and negative when the ranking is anti-correlated with model importance.
Occlusion units align exactly with \gls{vit} patch tokens. Full protocol details are given in
\cref{sec:srg}.

\Cref{tab:main_benchmark} reports \gls{srg} scores when explaining output logits. \Gls{ours}
achieves the highest faithfulness on six of seven models and is within one standard error of the
best on the seventh (ViT-L/14, where LeGrad leads by $0.05$), improving over \gls{attnlrp} by
$0.7$--$4.7$ \gls{srg} points. The gains are stable across model scales and pre-training regimes.
LeGrad is the strongest competitor, matching \gls{ours} on the CLIP-pretrained ViTs but degrading
on the remaining backbones, where it is either undefined or far behind.


\begin{table}[t]
  \centering
  \caption{Attribution quality, reported as \emph{\glsentryshort{srg} / Loc.}:
    \glsentryshort{srg} faithfulness (patch-wise occlusion, logit target) and localization, $n=500$ images each, higher is better. Best per
    metric and model in bold, second-best underlined. Standard errors are $\le 0.16$ and $\le 0.01$,
    full tables and additional backbones see
    \cref{tab:srg:logit:patch:339306,tab:localization:v2targetmask}. ``n.a.'': method not defined.}
  \label{tab:main_benchmark}
  \resizebox{\textwidth}{!}{%
    \begin{tabular}{lccccccc}
      \toprule
      \glsentryshort{srg} / Loc.
                                    & \textbf{ViT-B/16}
                                    & \textbf{ViT-L/14}
                                    & \textbf{ViT-H/14}
                                    & \textbf{DeiT3-L/16}
                                    & \textbf{DINOv2-L}
                                    & \textbf{SigLIP2-L/16}
                                    & \textbf{SwinV2-L}                                                                                                                                       \\
      \midrule

      \glsentryshort{ours}          & \textbf{3.20} / \textbf{0.62} & \underline{3.53} / \textbf{0.61} &
      \textbf{3.31} / \textbf{0.61} & \textbf{3.38} / 0.55          & \textbf{4.72} / \textbf{0.59}    &
      \textbf{5.26} / \textbf{0.51} & \textbf{2.36} / \textbf{0.64}                                                                                                                           \\ \glsentryshort{attnlrp} & 1.51 /
      0.32                          & 1.49 / 0.39                   & 1.65 / 0.41                      & 2.69 / 0.44             & 1.65 / 0.47 & 0.59 / 0.32             & \underline{0.93} /
      0.34                                                                                                                                                                                    \\ \glsentryshort{cplrp} & 1.33 / 0.36 & 1.23 / 0.46 & 1.33 / 0.41 & 2.97 / 0.54 & 0.36 / 0.44
                                    & 0.49 / 0.32                   & 0.80 / 0.32                                                                                                             \\ \midrule \glsentryshort{clrp} & 2.74 / \textbf{0.62} & 2.85 /
      \underline{0.51}              & 2.35 / 0.46                   & 2.55 / \underline{0.69}          & 4.43 / \underline{0.58} & 2.69 / 0.29 &
      n.a.                                                                                                                                                                                    \\ CheferAttnRollout & 2.74 / 0.50 & 2.85 / 0.46 & 2.35 / 0.44 & 2.65 / 0.61 &
      \underline{4.44} / 0.51       & n.a.                          & n.a.                                                                                                                    \\ GradAttnRollout & 2.46 / 0.50 & 2.84 / 0.49 & 2.43 /
      \underline{0.47}              & 2.87 / \textbf{0.70}          & 4.05 / 0.52                      & n.a.                    & n.a.                                                       \\ LibraFullGrad+ & 2.72 /
      \underline{0.54}              & 2.98 / 0.49                   & 2.10 / 0.43                      & 3.07 / 0.54             & 3.18 / 0.48 & 3.56 / \underline{0.37}
                                    & n.a.                                                                                                                                                    \\ LeGrad & \underline{3.18} / 0.53 & \textbf{3.58} / 0.47 & \underline{3.16} / 0.46 &
      \underline{3.12} / 0.51       & n.a.                          & \underline{4.20} / 0.33          & n.a.                                                                                 \\ \midrule \glsentryshort{ig} &
      1.23 / 0.45                   & 1.36 / 0.38                   & 1.19 / 0.31                      & 1.10 / 0.30             & 1.24 / 0.43 & 1.38 / 0.27             & 0.57 / 0.35        \\
      \glsentryshort{ixg}           & 0.45 / 0.40                   & 0.33 / 0.26                      & 0.27 / 0.22             & 0.23 / 0.25 & 0.45 / 0.42             & 0.27 /
      0.22                          & 0.43 / 0.31                                                                                                                                             \\ \midrule Random & -0.11 / 0.37 & -0.03 / 0.34 & -0.04 / 0.34 & -0.05 / 0.34 &
      -0.10 / 0.40                  & -0.04 / 0.29                  & 0.02 / \underline{0.39}                                                                                                 \\ \bottomrule
    \end{tabular}%
  }
\end{table}

\subsubsection{Localization: Attribution Localization}
While \gls{srg} evaluates whether an attribution correctly orders regions according to their importance to the model, localization captures a complementary, human-aligned notion of attribution quality: whether attribution mass is concentrated on semantically relevant regions, as identified by pixel-level ground-truth segmentation masks from ImageNet-S~\cite{gao2022luss}, independent of model behaviour.
The score is the fraction of total absolute attribution mass that falls within the
ground-truth object mask: a score of $1$ means all attribution is inside the object, and a score
equal to the relative mask area is the expected value for a spatially uniform attribution map. We
evaluate on 500 correctly classified images across 50 randomly sampled classes; restricting to
correctly classified samples ensures that attributions and segmentation masks refer to the same
object. Full details are given in \cref{sec:localization}.

\Cref{tab:main_benchmark} reports attribution localization scores. \Gls{ours} achieves the highest
or tied-highest localization on six of seven models, with scores of 0.51--0.64 compared to
0.32--0.47 for \gls{attnlrp}. The exception is DeiT3-L/16, where the rollout variants lead. On
ViT-B/16, \gls{attnlrp} even falls below the random baseline (0.32 vs.\ 0.37), consistent with the
relevance explosion analysis in \cref{fig:main_motivation}.



\subsubsection{{Ground-Truth Evaluation on FunnyBirds}}
\label{sec:funnybirds}

\begin{wraptable}[13]{r}{0.6\textwidth}
  \vspace{-\intextsep}
  \centering
  \caption{FunnyBirds results using the official dataset, metric implementations, and
    ViT-B/16 checkpoint of \citet{hesse2023funnybirds}. Higher is better for all metrics.}
  \label{tab:funnybirds}
  \resizebox{0.6\textwidth}{!}{
    \begin{tabular}{lcccccc}
      \toprule
      Method
                           & CSDC
                           & PC
                           & DC
                           & Distr.\
                           & SD
                           & TS                                                              \\
      \midrule
      \glsentryshort{ours} (ours)
                           & \textbf{0.942}
                           & \textbf{0.954}
                           & \textbf{0.916}
                           & \textbf{0.937}
                           & \textbf{0.774}
                           & 0.950                                                           \\
      \glsentryshort{attnlrp}
                           & 0.926
                           & 0.926
                           & 0.892
                           & 0.864
                           & 0.682
                           & 0.955                                                           \\
      \glsentryshort{cplrp}
                           & 0.803
                           & 0.620
                           & 0.610
                           & 0.440
                           & 0.437
                           & 0.831                                                           \\
      \glsentryshort{ixg}
                           & 0.743
                           & 0.584
                           & 0.602
                           & 0.432
                           & 0.510
                           & 0.669                                                           \\
      \glsentryshort{ig}
                           & 0.891
                           & 0.864
                           & 0.850
                           & 0.902
                           & 0.652
                           & 0.911                                                           \\
      Rollout
                           & 0.856
                           & 0.804
                           & 0.820
                           & 0.797
                           & 0.756
                           & 0.000                                                           \\
      \glsentryshort{clrp} & 0.916          & 0.922 & 0.898 & 0.899 & 0.733 & \textbf{0.957} \\ \bottomrule
    \end{tabular}
  }
\end{wraptable}

\gls{srg} measures faithfulness via occlusion and ImageNet-S localization measures spatial
plausibility. Neither compares against known ground truth of what the model relies on. We therefore
evaluate on the full FunnyBirds framework~\cite{hesse2023funnybirds}, a synthetic benchmark in
which every bird consists of discrete parts that can be removed and re-rendered in-distribution.
The single-deletion (SD) score directly correlates attributed part importance with ground-truth
causal importance obtained by intervention. As shown in \cref{tab:funnybirds}, \gls{ours} achieves
the best score on all four completeness metrics and on correctness, with an SD of 0.774 versus
0.682 for \gls{attnlrp}, a +13\% improvement, while on par with the best methods on contrastivity.
Our three evidence types thus agree. Occlusion faithfulness, localization, and controlled causal
ground truth all favour \gls{ours}. {Protocol details are given in \cref{app:funnybirds}.}

\subsection{Isolating Residual Cancellation as the Cause}
\label{sec:causal}

The benchmarks above show that \gls{ours} improves attribution quality, but not yet that the
improvement is driven by correcting residual cancellation rather than a generic regularization
effect of the $\gamma$-rule. We establish this causally with two analyses.

\paragraph{Targeted channel-wise interventions.}
We apply the residual $\gamma$-rule only to selected channels at every residual merge. If the gains
stemmed from generic regularization, treating more channels should help regardless of which ones
are selected. Instead, \cref{tab:interventions} shows the opposite on ViT-B/16. Correcting only the
top 0.5\% most cancellation-prone channels (92) improves \gls{srg} by +7.9\% over \gls{attnlrp},
whereas correcting 75\% of low-cancellation channels (13{,}824, a $150\times$ larger set) decreases
\gls{srg} by $-5.5$\%. Across all interventions, \gls{srg} correlates much more strongly with the
amount of unstable relevance removed (Spearman $\rho = 0.79$) than with the number of modified
channels ($\rho = 0.50$). The pattern replicates on DINOv2-B (\cref{app:causal_details}). The gains
of \gls{ours} therefore arise specifically from correcting residual cancellation.

\paragraph{Component ablations.}
We disable the \gls{attnlrp} attention and LayerNorm rules to assess their contribution relative to the residual $\gamma$-rule. While both components contribute to attribution quality, \gls{ours} is substantially more robust to their removal: ablating either rule causes a much smaller drop in \gls{srg} than the corresponding performance gap between \gls{ours} and \gls{attnlrp}. For example, on SigLIP2, \gls{ours} achieves an \gls{srg} of 3.27 compared to $-0.004$ for \gls{attnlrp} when the attention rule is removed. This indicates that the residual $\gamma$-rule itself captures substantially more of the stability and faithfulness measured by \gls{srg} than either the attention or LayerNorm rule individually, while the latter provide additional gains. Full tables are given in \cref{app:causal_details}.


\subsection{Localizing SAE Features in Input Space}
\label{sec:sae-attribution}

\begin{figure}[t]
  \centering
  \includegraphics[width=1.0\linewidth]{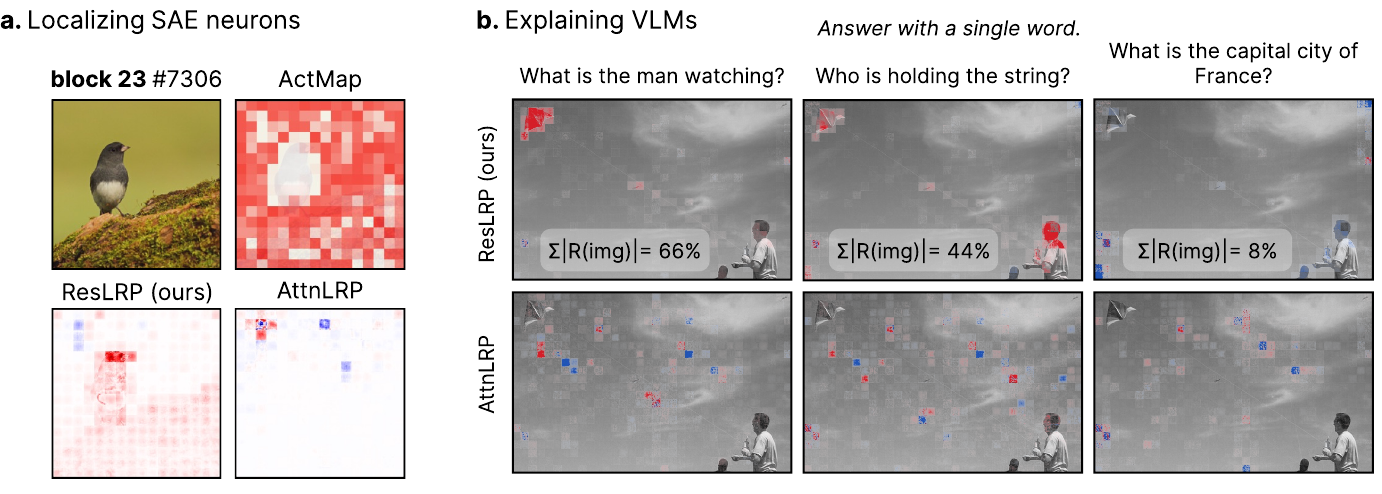}
  \caption{{Qualitative Examples.}
    \textbf{a)} \Gls{sae} activation maps (\mbox{ActMap}) need not align with the encoded concept
    (``bird''), whereas \gls{ours} highlights inputs faithful to the latent (cf.\ \cref{tab:sae_srg}).
    \textbf{b)} \Gls{ours} grounds \gls{vlm} text generation (Qwen3-VL-4B-Instruct) at pixel level,
    per output token (\cref{sec:vlm}).}
  \label{fig:qualitative}
\end{figure}

\Glspl{sae} decompose activations into sparse, potentially interpretable features and have become a
central tool in mechanistic interpretability~\cite{cunningham2023sparse,templeton2024scaling}.
Latents are conventionally visualized by reshaping per-token \gls{sae} activations to the patch
grid (ActMap)~\cite{lim2025sparse}. This can mislead: self-attention routes information across
tokens before the \gls{sae} hook point, so the token with the highest activation need not
correspond to the image region that caused it. Input attribution traces the full graph from pixels
to the latent and recovers this link. We train TopK-\glspl{sae}~\cite{gao2025scaling} at three
blocks of a frozen ViT-L/14 backbone, attribute the summed post-TopK activation of individual
latents back to the input, and measure faithfulness with the \gls{srg} protocol using the latent
activation in place of the classifier logit (full setup in \cref{sec:sae_details}). \Gls{ours}
attains the best \gls{srg} and the best mean rank at all three blocks (\cref{tab:sae_srg}), e.g.\
$0.449 \pm 0.006$ versus $0.334 \pm 0.006$ for \gls{attnlrp} at \texttt{blocks.15}. The ActMap
baseline collapses to $-0.031 \pm 0.006$ at \texttt{blocks.23}, i.e.\ slightly anti-correlated with
the very latent it is meant to visualize, confirming that activation maps lose spatial meaning at
depth. \Cref{fig:qualitative}a shows the same effect qualitatively.

\subsection{Faithful Attributions for Large Vision-Language Models}
\label{sec:vlm-main} \label{sec:vlm}

{ \Gls{attnlrp} is, to our knowledge, one of the most widely deployed attribution methods for
  \glspl{vlm}~\citep{biton2026hidden,gong2026saliency}, yet these composite architectures are deeper
  than any \gls{vit} evaluated above. We therefore repeat both evaluations on three instruction-tuned
  \glspl{vlm} --- Qwen2.5-VL-3B, Qwen3-VL-4B and Gemma-3-4B-it --- attributing the logit of a single
  answer token with respect to the input pixels. \Cref{tab:vlm} reports the results. \Gls{ours} is
  best on both metrics on all three models. Localization improves over \gls{attnlrp} by $+0.09$ to
  $+0.10$ absolute ($+27\%$ to $+29\%$ relative), and \gls{srg} improves by a factor of $1.9$ to
  $3.4$. Every gap is significant under a two-sided Wilcoxon signed-rank test over per-image scores
  ($p \le 1.7\times10^{-26}$). Two observations sharpen the picture. First, the relative localization
  gain is remarkably uniform across three unrelated vision towers and two tokenisation schemes, which
  is what a mechanism rooted in the residual stream, rather than in any architectural particular,
  predicts. Second, on Gemma-3-4B-it \gls{attnlrp} does not separate from \gls{ixg} on localization
  at all ($0.352$ vs.\ $0.352$) and barely does on \gls{srg} ($+0.21$ vs.\ $+0.17$), the
  $\gamma$-rule on the residual merges is what recovers a usable map. This mirrors the ViT-B/16~A
  case in \cref{tab:main_benchmark}, where \gls{attnlrp} falls below the random baseline,
  and supports the same reading: once cancellation is severe enough, \gls{attnlrp}'s advantage over a
  plain gradient explanation disappears, and correcting the residual merge restores it. Because the
  three models differ in image tokenisation, and therefore in the natural occlusion unit, \gls{srg}
  magnitudes are comparable within a row but not down a column. The full protocol, including prompt
  design, sample selection and the per-model occlusion geometry, is given in \cref{sec:vlm_eval}. }

\begin{table}[h]
  \centering
  \caption{\textbf{Attribution quality on vision-language models.} Attribution localization
    (\cref{eq:loc}) and \gls{srg} (\cref{eq:srg}) as \emph{Loc.\ / \gls{srg}}, higher is better,
    mean $\pm$ one standard error, $n=500$.
    \Gls{srg} magnitudes
    are comparable within a row but not across rows, see \cref{sec:vlm_eval}.}
  \label{tab:vlm}
  \resizebox{\textwidth}{!}{%
    \begin{tabular}{lccc}
      \toprule
      Loc.\ / \gls{srg} & \glsentryshort{ixg}              & \glsentryshort{attnlrp}          & \glsentryshort{ours}                        \\
      \midrule
      Qwen2.5-VL-3B     & 0.256$\pm$0.012 / +0.08$\pm$0.04 & 0.365$\pm$0.011 / +1.29$\pm$0.12 & \textbf{0.464$\pm$0.011 / +2.39$\pm$0.17  } \\
      Qwen3-VL-4B-it    & 0.274$\pm$0.013 / +0.13$\pm$0.11 & 0.325$\pm$0.012 / +3.87$\pm$0.20 & \textbf{0.418$\pm$0.010 / +13.04$\pm$0.30 } \\
      Gemma-3-4B-it     & 0.352$\pm$0.013 / +0.17$\pm$0.06 & 0.352$\pm$0.014 / +0.21$\pm$0.04 & \textbf{0.455$\pm$0.012 / +0.41$\pm$0.03  } \\
      \bottomrule
    \end{tabular}%
  }
\end{table}

\section{Conclusion}
We identify a previously overlooked failure mode in \gls{lrp}-based explanation of \glspl{vit},
namely attribution instability driven by residual cancellations. Destructive interference in the
forward pass makes state-of-the-art \gls{lrp} variants divide by near-zero activations, amplifying
relevance into noisy, contradictory attributions that overshadow more stable signals, an effect
most pronounced in early \gls{vit} layers and far weaker in language models. We address it with
\gls{ours}, a bounded, $\gamma$-stabilized redistribution rule for residual additions that is
exactly conservative and prevents local amplification from cascading through the network. Occlusion
faithfulness, localization, and controlled ground truth on FunnyBirds all favour \gls{ours}, and
targeted interventions confirm the gains come specifically from correcting residual cancellation.
The largest gains appear in modern \glspl{vlm}, where prior methods are either too noisy or too
expensive to apply, opening follow-up work from bias detection to attribution-guided training for
visual grounding.

\paragraph{Limitations \& Outlook.}
The single hyperparameter $\gamma$ shows a broad performance plateau with the untuned default
$\gamma = 1$ near-optimal across all evaluated models (\cref{sec:gamma}), and \gls{ours} improves
faithfulness and stability together rather than trading one for the other. Our guarantees are local
to the residual additions, so full-model attribution quality rests on the three agreeing empirical
evidence types. End-to-end conservation of the full pipeline is limited by the \gls{attnlrp}
attention rule, and exact conservation is available by combining \gls{ours} with the \gls{cplrp}
attention rule at a known quality cost. We do not examine whether similar cancellations arise in
other architectures, such as diffusion transformers or state-space models, where residual streams
and update branches may interact differently, and note that strong residual cancellation may have
broader implications for training dynamics and generalization. Finally, a better understanding of
attribution failures could also be used to design models that produce convincing yet misleading
explanations.

\section*{Acknowledgments}
This work was supported by
the Federal Ministry of Research, Technology and Space (BMFTR) as grants [BIFOLD (01IS18025A, 01IS180371I), xJuRAG (16IS25015B)];
the European Union’s Horizon Europe research and innovation programme (EU Horizon Europe) as grant ACHILLES (101189689);
and the German Research Foundation (DFG) as research unit DeSBi [KI-FOR 5363] (459422098).

\bibliographystyle{unsrtnat} \bibliography{main}

\appendix \setcounter{table}{0} \setcounter{figure}{0}
\renewcommand{\thetable}{\Alph{section}.\arabic{table}}
\renewcommand{\thefigure}{\Alph{section}.\arabic{figure}}
\section{Technical Appendices and Supplementary Material}
\label{sec:appendix}

{This appendix collects the technical material supporting the main text, grouped into analysis
  and theory, experimental setup, metric definitions, protocols for the component-level and
  multimodal experiments, and extended results.}

\begin{itemize}
  \item {\textbf{Analysis and theory.} \Cref{app:cancellation} measures residual cancellation and
          relevance amplification across \glspl{vit} and \glspl{llm}. \Cref{app:proof} proves the
          boundedness proposition stated in \cref{methods:boundness}. \Cref{app:implementation} gives the
          PyTorch implementation of the residual $\gamma$-rule. \Cref{sec:resilrp-plus} describes
          \gls{ours}$+$, the composition with gradient decomposition that removes patch-grid artifacts.}
  \item {\textbf{Experimental setup.} \Cref{sec:setup} lists the model checkpoints, datasets,
          libraries, and hardware used for all benchmarks. \Cref{app:baselines} gives formal descriptions
          of the baseline attribution methods.}
  \item {\textbf{Metrics.} \Cref{sec:srg} formalizes the \textit{Symmetric Relevance Gain}
          (\gls{srg}) faithfulness metric, including the MIF/LIF occlusion protocol and tie-breaking.
          \Cref{sec:localization} describes the \textit{Attribution Localization} score used to evaluate
          spatial grounding against ground-truth segmentations. \Cref{app:funnybirds} details the
          FunnyBirds protocol and the six reported ground-truth metrics.}
  \item {\textbf{Component-level and multimodal protocols.} \Cref{sec:sae_details} specifies the
          architecture, hyperparameters, and evaluation protocol for the \glspl{sae}.
          \Cref{sec:vlm_eval} details the forced-choice localization design, the explained scalar, and the
          per-model occlusion geometry used for the \glspl{vlm}.}
  \item {\textbf{Extended results.} \Cref{app:extended_benchmarks} reports the full per-model
          faithfulness and localization tables shortened in \cref{tab:main_benchmark}. \Cref{sec:gamma}
          sweeps the $\gamma$ parameter. \Cref{sec:registers} shows that register tokens reduce residual
          amplification. \Cref{app:causal_details} covers the intervention protocol, the replication on
          DINOv2-B, and the component ablations. \Cref{app:qualitative_examples} collects qualitative
          attribution examples.}
\end{itemize}

\subsection{Residual Cancellations}
\label{app:cancellation}

Residual cancellations are a primary source of relevance amplification in \glspl{vit}. We evaluate
a diverse set of \glspl{vit} and autoregressive LLMs of comparable parameter scale, including
standard ImageNet classifiers, CLIP-based vision transformers, and instruction-tuned language
models.

\begin{figure}[t]
  \centering
  \includegraphics[width=0.98\linewidth]{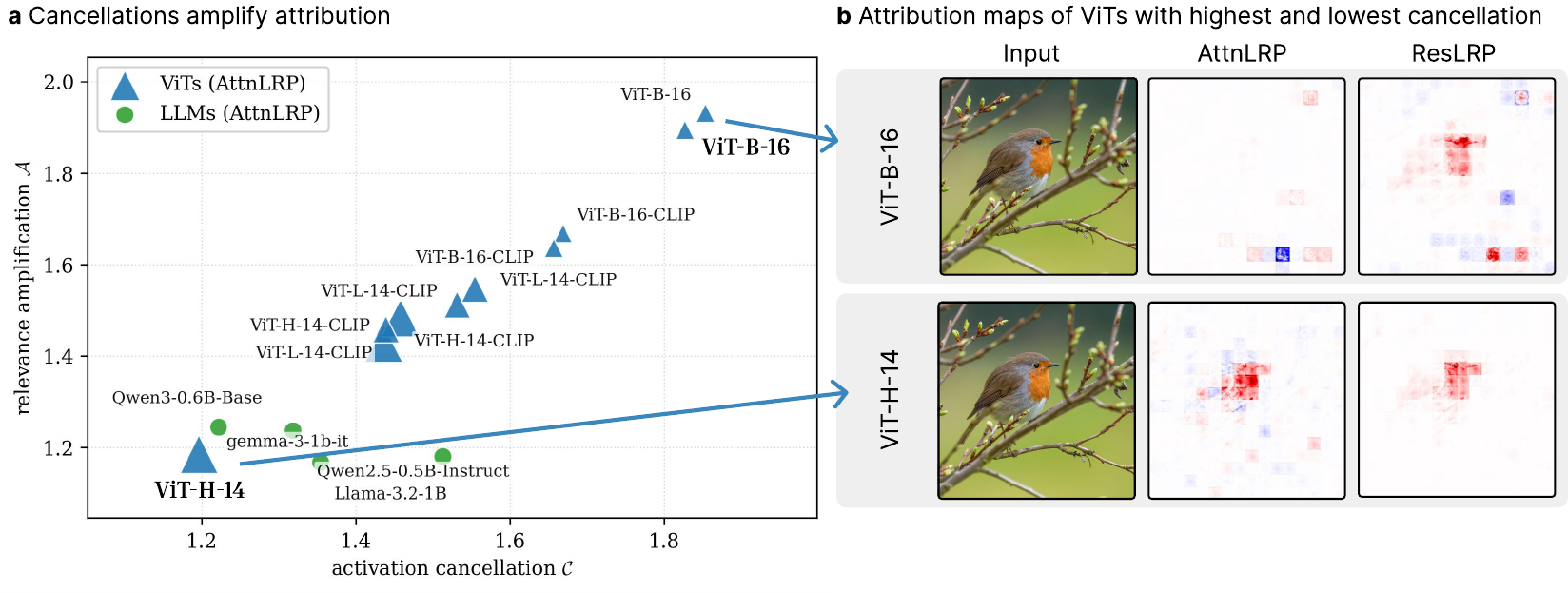}
  \caption{Residual cancellation correlates with relevance amplification in ViTs. \textbf{(a)} Across a diverse set of ViTs and LLMs, residual cancellation exhibits a near-perfect linear correlation with relevance amplification. Marker size corresponds to model parameter count. Larger models generally show reduced cancellation, while CLIP-based ViTs tend to exhibit substantially less effects than standard ImageNet-pretrained models. \textbf{(b)} Corresponding \gls{attnlrp} heatmaps for the best ViT (\texttt{vit\_huge\_patch14\_224.orig\_in21k}) and worst ViT (\texttt{vit\_base\_patch16\_224.augreg\_in21k}) demonstrate that strong cancellation leads to noisy and diffuse explanations, whereas reduced cancellation yields cleaner and more localized relevance maps. The artifacts of \gls{ours} heatmap for ViT-B-16 emerge from register tokens~\cite{Darcet2023VisionTN}.}
  \label{app:fig:main_motivation}
\end{figure}

In Figure~\ref{app:fig:main_motivation}\textbf{(a)} we observe a near-perfect linear relationship
between the degree of residual cancellation and relevance amplification, indicating that stronger
cancellation effects directly lead to amplified attribution noise. Across both modalities, larger
models generally exhibit weaker cancellation effects. Among the evaluated ViTs,
\texttt{vit\_huge\_patch14\_224.orig\_in21k} shows the lowest cancellation despite substantially
less training than the CLIP-based variants, whereas \texttt{vit\_base\_patch16\_224.augreg\_in21k}
exhibits the strongest cancellation. Overall, CLIP-trained models tend to suffer from significantly
higher cancellation compared to standard ImageNet-pretrained models.

In Figure~\ref{app:fig:main_motivation}\textbf{(b)} qualitative heatmaps further illustrate this
effect. For the model with the worst cancellation score, \gls{attnlrp} produces highly noisy and
spatially diffuse explanations. In contrast, the best-performing ViT yields substantially cleaner
and more localized relevance maps, demonstrating that reduced residual cancellation directly
improves attribution quality.

{\subsection{Proof of the Boundedness Proposition}
  \label{app:proof} We begin by noting the symmetry
  \begin{equation}
    \label{eq:sym}
    \Phi(a,b) + \Phi(b,a) = 1,
  \end{equation}
  which implies $\Phi(b,a) = 1 - \Phi(a,b)$.
  Since the interval $[-\tfrac{1}{\gamma},\,1+\tfrac{1}{\gamma}]$ is invariant under
  $x \mapsto 1-x$, it suffices to prove the bound for the case $|a| \ge |b|$.
  (The degenerate case $|a|=|b|$ with opposing signs is excluded by $c \neq 0$.)

  We distinguish two cases.

  \emph{Case 1: $\operatorname{sign}(a) = \operatorname{sign}(b)$.} Both signs agree with
  $\operatorname{sign}(c)$, so both indicators equal $1$ and the factor $(1+\gamma)$ cancels. We
  obtain
  \[
    \Phi(a,b) = \frac{a}{a+b} \in [0,1]
    \subseteq \left[-\tfrac{1}{\gamma},\,1+\tfrac{1}{\gamma}\right].
  \]

  \emph{Case 2: $\operatorname{sign}(a) \neq \operatorname{sign}(b)$ and $|a| > |b|$.} Then
  $\operatorname{sign}(c) = \operatorname{sign}(a)$, so the first indicator is $1$ and the second is
  $0$. Writing $b = \alpha a$ with $\alpha \in (-1,0]$, we obtain
  \[
    \Phi(a,b)
    =
    \frac{(1+\gamma)a}{(1+\gamma)a + b}
    =
    \frac{1+\gamma}{1+\gamma+\alpha}.
  \]
  Since $\alpha \in (-1,0]$, the denominator satisfies $\gamma < 1+\gamma+\alpha \le 1+\gamma$,
  yielding
  \[
    1 \le \Phi(a,b) \le \frac{1+\gamma}{\gamma}
    = 1+\frac{1}{\gamma}.
  \]





  This establishes the bound for $|a| \ge |b|$, the remaining case follows from the symmetry in
  \cref{eq:sym}. \newline\phantom{.}\hfill$\square$}

\subsection{Implementation}
\label{app:implementation}

\gls{ours} is implemented on top of \gls{attnlrp} with straight-through-estimator style
\texttt{.detach()} tricks, so the forward pass is unchanged and only gradients are modified. The
residual $\gamma$-rule of \cref{eq:resilrp} adds a single backward hook per residual addition, as
shown in \cref{tab:lrp-tricks}.
We use \texttt{zennit v1.0}~\cite{anders2021software} (GNU Lesser General Public License v3 or
later) and \texttt{LXT v2.1}~\cite{achtibat2024attnlrp} (BSD 3-Clause License) for computing LRP
attributions.
\begin{table}[t]
  \centering
  \caption{PyTorch implementation of \glsentryshort{lrp} using techniques inspired by straight-through estimators: the forward pass is unchanged, while \texttt{.detach()} modifies gradients to enforce \glsentryshort{lrp}-style gradients. The $\gamma$-extension on the residual stream is applied via a backward hook. Finally, the input heatmap can be computed as the element-wise product of input embeddings and their gradients.}
  \label{tab:lrp-tricks}
  \begin{tcolorbox}[
      enhanced, boxrule=0pt, frame hidden, colback=boxbg,
      arc=4pt, boxsep=2pt,
      left=4pt, right=4pt, top=2pt, bottom=2pt
    ]
    \small
    \centering
    \begin{tabular}{@{}ll@{}}
      \toprule
      \textbf{Operation}  & \textbf{PyTorch Implementation Trick}                                                \\
      \midrule
      \multicolumn{2}{@{}l}{\textcolor{codegray}{\textbf{\textit{Standard \glsentryshort{attnlrp} Operations}}}} \\
      LayerNorm           & \texttt{y = (x - x.mean()) / \detached{[}x.var().sqrt()\detached{].detach()}}        \\
      GELU                & \texttt{y = x * \detached{[}$\Phi$(x)\detached{].detach()}}                          \\
      Query-Key           & \texttt{y = 0.5 * (s := Q @ K) + \detached{[}0.5 * s\detached{].detach()}}           \\
      Attention           & \texttt{y = 0.5 * (z := A @ V) + \detached{[}0.5 * z\detached{].detach()}}           \\
      \midrule
      \multicolumn{2}{@{}l}{\textcolor{codegray}{\textbf{\textit{Proposed \glsentryshort{ours} Extension}}}}     \\
      Residual ($\gamma$) &
      \begin{tabular}[t]{@{}l@{}}
        \texttt{\textcolor{codegray}{\# Forward:} y = x\_in + x\_up}                                 \\
        \texttt{\textcolor{codegray}{\# Scaling:} w = lambda z: 1 + $\gamma$ * (z * y > 0)}          \\
        \texttt{\phantom{\# Scaling: }$\rho$ = lambda z: w(z) / (w(x\_in)*x\_in + w(x\_up)*x\_up)}   \\
        \texttt{\textcolor{codegray}{\# Hooks:  } x\_in.register\_hook(lambda g: g * $\rho$(x\_in))} \\
        \texttt{\phantom{\# Scaling: }x\_up.register\_hook(lambda g: g * $\rho$(x\_up))}
      \end{tabular}             \\
      \bottomrule
    \end{tabular}
  \end{tcolorbox}
\end{table}

\subsection{\texorpdfstring{\gls{ours}$+$}{ResiLRP+}: Composition with Gradient Decomposition}
\label{sec:resilrp-plus}

\begin{figure}[b]
  \centering
  \includegraphics[width=\linewidth]{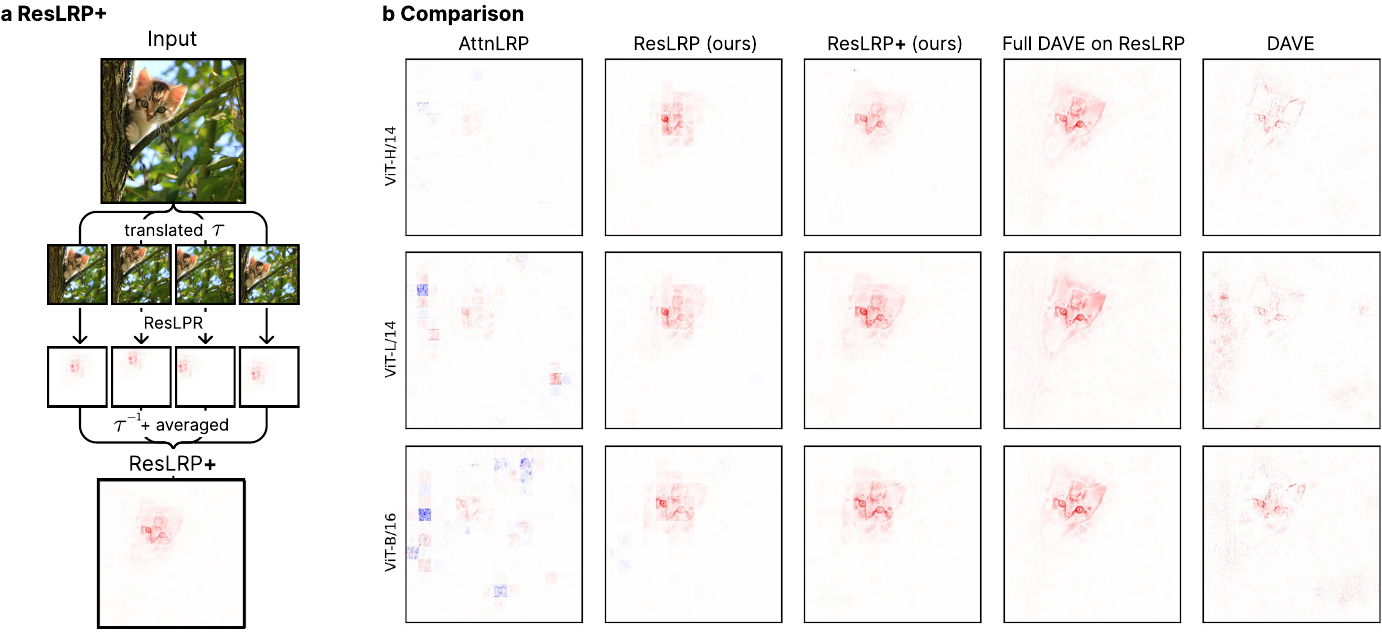}
  \caption{
    \textbf{Removing patch artifacts.}
    \gls*{ours} yields clearer, more localized attribution maps than prior methods, by reducing attribution noise. \gls*{ours}+ removes patch-grid artifacts by averaging attributions over four small spatial perturbations of the input, further improving localization and visual coherence.}
  \label{fig:qualitative:dave}
\end{figure}

\gls{ours} corrects relevance where it is redistributed across residual merges, along the backward
path. A complementary artifact arises at the very end of that path, where the patch embedding
projects relevance back to pixels and imposes a blocky, patch-aligned structure.
\gls{dave}~\cite{dave} addresses precisely this by decomposing the input gradient into locally
equivariant and artifact-induced components, using 50 forward passes over spatially translated inputs.
The two act on disjoint parts of the computation and
compose directly, applying \gls{dave}'s decomposition to the relevance \gls{ours} delivers at the
patch embedding instead of the plain input-times-gradient step. We find that already only \emph{four}
steps of spatial input translations without any additional transformation yield much reduced
patch-artifacts. In contrast to the 50 forward passes used by \gls{dave}, this requires only four
forward passes, making the resulting correction substantially more computationally efficient.
We denote this configuration \gls{ours}$+$. 
Figure~\ref{fig:qualitative:dave} illustrates this process and the resulting reduction in patch artifacts.


\subsection{Experimental Setup}
\label{sec:setup}

\paragraph{Models.}
We evaluate publicly available checkpoints spanning contrastive--multimodal (CLIP, SigLIP2),
supervised (DeiT3), self-supervised (DINOv2, with and without register tokens), and hierarchical
(SwinV2) pre-training, across patch sizes 14 and 16 and model scales from Small to Giant
(\cref{tab:models}). Checkpoints are loaded via TIMM~\cite{Wightman_PyTorch_Image_Models} or the
HuggingFace Hub as indicated. Covering several families and recipes ensures that findings are not
artefacts of a single training regime. Not every method is defined for every backbone, and not
every backbone enters every experiment, cells marked ``n.a.'' in the result tables indicate the
former.

\begin{table}[h]
  \centering
  \caption{{Evaluated models, grouped by pre-training family.
      ``(Reg.)'' denotes a variant trained with register tokens~\cite{Darcet2023VisionTN}.}}
  \label{tab:models}
  \small
  \begin{tabular}{ll}
    \toprule
    \textbf{Short name} & \textbf{Identifier}                                                     \\
    \midrule
    \multicolumn{2}{@{}l}{\textit{Contrastive (CLIP), fine-tuned on ImageNet}}                    \\
    ViT-B/16            & \texttt{timm/vit\_base\_patch16\_clip\_224.openai\_ft\_in12k\_in1k}     \\
    ViT-L/14            & \texttt{timm/vit\_large\_patch14\_clip\_224.laion2b\_ft\_in1k}          \\
    ViT-H/14            & \texttt{timm/vit\_huge\_patch14\_clip\_224.laion2b\_ft\_in1k}           \\
    \midrule
    \multicolumn{2}{@{}l}{\textit{Contrastive multimodal}}                                        \\
    SigLIP2-L/16        & \texttt{google/siglip2-large-patch16-256}                               \\
    \midrule
    \multicolumn{2}{@{}l}{\textit{Supervised}}                                                    \\
    {DeiT3-M/16}        & \texttt{timm/deit3\_medium\_patch16\_224.fb\_in1k}                      \\
    {DeiT3-L/16}        & \texttt{timm/deit3\_large\_patch16\_224.fb\_in22k\_ft\_in1k}            \\
    \midrule
    \multicolumn{2}{@{}l}{\textit{Hierarchical}}                                                  \\
    SwinV2-L            & \texttt{timm/swinv2\_large\_window12to16\_192to256.ms\_in22k\_ft\_in1k} \\
    \midrule
    \multicolumn{2}{@{}l}{\textit{Self-supervised (DINOv2), linear ImageNet-1k head}}             \\
    DINOv2-S            & \texttt{facebook/dinov2-small-imagenet1k-1-layer}                       \\
    DINOv2-S (Reg.)     & \texttt{facebook/dinov2-with-registers-small-imagenet1k-1-layer}        \\
    {DINOv2-B}          & \texttt{facebook/dinov2-base-imagenet1k-1-layer}                        \\
    {DINOv2-B (Reg.)}   & \texttt{facebook/dinov2-with-registers-base-imagenet1k-1-layer}         \\
    DINOv2-L            & \texttt{facebook/dinov2-large-imagenet1k-1-layer}                       \\
    DINOv2-L (Reg.)     & \texttt{facebook/dinov2-with-registers-large-imagenet1k-1-layer}        \\
    DINOv2-G            & \texttt{facebook/dinov2-giant-imagenet1k-1-layer}                       \\
    DINOv2-G (Reg.)     & \texttt{facebook/dinov2-with-registers-giant-imagenet1k-1-layer}
    \\ \bottomrule
  \end{tabular}
\end{table}

\paragraph{Datasets.}
The two metrics use different datasets suited to their evaluation protocol. \emph{\gls{srg}} uses a
stratified random subset of the ImageNet-1k validation split~\cite{Deng2009imagenet}: 500~images
sampled with fixed seed~339306. \emph{Localization} uses ImageNet-S~\citep{gao2022luss}:
50~randomly selected classes (seed~754068), 10~correctly classified samples per class, for a total
of 500~images (see Section~\ref{sec:localization} for the sample selection rationale). In both
cases each image is preprocessed with the model-specific TIMM transform (resize, centre-crop,
normalisation).

\paragraph{LRP libraries.}
We use \texttt{zennit v1.0}~\cite{anders2021software} (GNU Lesser General Public License v3 or
later) and \texttt{LXT v2.1}~\cite{achtibat2024attnlrp} (BSD 3-Clause License) for computing LRP
attributions.

\subsection{Description of Baseline Attribution methods}
\label{app:baselines}
\subparagraph{\gls{ixg}}

\gls{ixg} is one of the most intuitive methods to determine how sensitive a trained model is to its
input features. By multiplying the gradient by the corresponding input features, the method
produces a local linear approximation of the model at the given input~\cite{Simonyan14a}:

\begin{equation}
  \text{I$\times$G}(\mathbf{x}) = \frac{\partial f_c(\mathbf{x})}{\partial \mathbf{x}} \times \mathbf{x}
\end{equation}

where $f_c(\mathbf{x})$ is the model's output for class $c$ given input $\mathbf{x}$.

This method is known to suffer from the \emph{gradient shattering} effect \cite{balduzzi17}, which
produces very noisy heatmaps, especially in ReLU-based \glspl{cnn}.

\subparagraph{\gls{ig}}
In order to reduce noise in the \gls{ixg} method, the \gls{ig} method integrates gradients along a
trajectory from a baseline $\mathbf{x'}$ to the input $\mathbf{x}$, approximated by \emph{m}
interpolation steps \cite{sundararajan17}:

\begin{equation}
  \begin{aligned}
    \text{IG}(\mathbf{x}) & = (\mathbf{x} - \mathbf{x}') \int_{\alpha=0}^{1} \frac{\partial f_j(\mathbf{x}' + \alpha \times (\mathbf{x} - \mathbf{x}'))}{\partial \mathbf{x}} \, d\mathbf{x}                        \\
                          & \approx (\mathbf{x} - \mathbf{x}') \sum_{k=1}^{m} \frac{\partial f_j\!\left(\mathbf{x}' + \frac{k}{m} \times (\mathbf{x} - \mathbf{x}')\right)}{\partial \mathbf{x}} \times \frac{1}{m}
  \end{aligned}
\end{equation}

\subparagraph{\gls{legrad}}

\gls{legrad} computes the gradient of the model's output with respect to the attention maps of
individual \gls{vit} layers, using the gradient itself as the explainability signal. The method
aggregates this signal across all layers, combining activations from both the intermediate and
final tokens to produce a unified explainability map~\cite{bousselham25}.

\subparagraph{\gls{fullgrad}}
\gls{fullgrad} is an extension of \gls{ixg}: it not only includes input features but also bias terms at each layer. The \gls{fullgrad} attribution map is calculated as \cite{srinivas19}:

\begin{equation}
  \text{FullGrad}(\mathbf{x}) = \text{I$\times$G}(f_c)(\mathbf{x}) + \sum_{l=1}^{L} \sum_{b \in \mathcal{B}_l} \text{I$\times$G}(f_c^{(b)})(b)
\end{equation}

where $\text{I$\times$G}(f_c^{(b)})(b)$ is the Input $\times$ Gradient attribution map of the
sub-network $f_c^{(b)}$ with a bias term $b$ from layer $l$ as input. Here, $f_c^{(b)}$ denotes the
sub-network of $f_c$ starting from bias term $b$ until the output, and $\mathcal{B}_l$ denotes the
set of all bias terms in layer $l$.

\subparagraph{FullGrad+}
FullGrad+ builds on \gls{fullgrad} by incorporating the PLUS technique, which aggregates
attribution maps for input and bias terms in every layer \cite{mehri24}:

\begin{equation}
  \text{FullGrad+}(\mathbf{x}) = \sum_{l=1}^{L} \text{I$\times$G}(f_l)(\mathbf{x}_l) + \sum_{l=1}^{L} \sum_{b \in \mathcal{B}_l} \text{I$\times$G}(f_b)(b)
\end{equation}

where $f_l$ denotes the sub-network from layer $l$ to the output, $\mathbf{x}_l$ is the
intermediate activation at layer $l$, and $f_b$ denotes the sub-network from bias term $b$ to the
output.

\subparagraph{Libra FullGrad+}

Libra FullGrad+ applies the LibraGrad framework, which aims at restoring balanced gradients, to
FullGrad+. LibraGrad restores FullGrad-completeness (FG-completeness), a property ensuring
attributions faithfully decompose model outputs, which modern Transformers violate due to
non-locally-affine operations. It does so by pruning and scaling backward paths without modifying
the forward pass \cite{mehri25}.

\subparagraph{Attention Rollout and Gradient Attention Rollout}
Attention Rollout addresses the unreliability of raw attention weights in higher Transformer
layers, where embeddings are mixtures of multiple input tokens. The method recursively multiplies
attention matrices across layers to trace information back to the input. The attention rollout is
computed as~\cite{abnar20}:

\begin{equation}
  \tilde{A}(l_i) = \begin{cases} A(l_i) \tilde{A}(l_{i-1}) & \text{if } i > j  \\ A(l_i) & \text{if } i = j \end{cases} \end{equation}

where $\tilde{A}(l_i)$ denotes the attention rollout at layer $l_i$, $A$ denotes raw attention, $j$
denotes the starting layer of the rollout, and the multiplication is a matrix multiplication.

Gradient Attention Rollout \cite{gildenblat2020} extends Attention Rollout by weighting each
attention map by the gradient of the target class output, making it class-specific, and then
averaging over the attention heads while masking out negative attentions.

\paragraph{Implementation details.}
Attributions are precomputed once per (model, method) pair and cached to disk; both \gls{srg} and
localization evaluation load the cached tensors, avoiding redundant attribution computation. For
\gls{srg}, occlusion curves use $T = 100$ evaluation steps per sample with the patch groups
combined linearly where the total patch count exceeds this value; batch sizes are 64 for
attribution computation, 32 for \gls{srg} evaluation (Base/Large), and 1 for Huge models. For
localization, attribution maps are channel-summed and resized to the segmentation mask resolution
when needed; the Quantus \texttt{AttributionLocalisation} metric is applied sample-by-sample with
batch size~8 except for Huge models where we use again a batch size of 1. All experiments run on a
single NVIDIA GeForce RTX 5090 with \texttt{float32} precision.

\subsection{Faithfulness Metric: Symmetric Relevance Gain}
\label{sec:srg}

A faithful attribution map should assign high scores to input regions the model genuinely relies
on. We measure faithfulness via the \emph{Symmetric Relevance Gain}
(\gls{srg})~\cite{blucher2024decoupling}, an occlusion-based benchmark that requires no
human-annotated ground truth and directly probes the model's own decision process.

\paragraph{MIF and LIF occlusion curves.}
Let $\mathbf{x} \in \mathbb{R}^{C \times H \times W}$ be an input image and $A \in \mathbb{R}^{H
    \times W}$ the corresponding attribution map (channel-summed if the method produces a multi-channel
output). The image is divided into $N$ non-overlapping patches $\mathcal{P} = \{p_1, \ldots,
  p_N\}$; each patch $p_k$ receives an importance score $s_k = \sum_{(i,j) \in p_k} A_{ij}$. Two
orderings are derived from these scores:

\begin{itemize}
  \item \textbf{Most-Important-First (MIF):} patches sorted by $s_k$ in descending order.
  \item \textbf{Least-Important-First (LIF):} patches sorted by $s_k$ in ascending order.
\end{itemize}

Starting from the original image, patches are progressively replaced by a fixed background value
(the per-channel ImageNet mean). Let $\mathbf{x}^{(t)}_{\text{MIF}}$ denote the image after
replacing the $t$ highest-scored patches, and let $z_y(\mathbf{x})$ be the model's output logit for
the true class $y$. The \emph{MIF curve} records the logit at each step:
\begin{equation}
  c_{\text{MIF}}(t) \;=\; z_y\!\left(\mathbf{x}^{(t)}_{\text{MIF}}\right),
  \quad t = 0, 1, \ldots, T,
\end{equation}
where $t = 0$ is the unoccluded input and $t = T$ is the fully
occluded (background-only) baseline.
The \emph{LIF curve} $c_{\text{LIF}}(t)$ is defined analogously,
removing the least-scored patches first.

\paragraph{\gls{srg} score.} For a faithful method the MIF curve should fall steeply (removing important patches hurts
quickly), while the LIF curve should remain high for longer (removing unimportant patches has
little effect). This divergence is summarised by the Symmetric Relevance Gain:
\begin{equation}
  \label{eq:srg}
  \text{\gls{srg}}
  \;=\; \frac{\mathrm{AUC}(c_{\text{LIF}}) \;-\; \mathrm{AUC}(c_{\text{MIF}})}{n},
\end{equation}
where AUC denotes the area under the curve computed via the trapezoidal
rule and $n$ is the number of evaluation steps (normalisation constant).
\gls{srg} is positive for a faithful method, zero for a random one, and
negative for a method anti-correlated with model importance.

\paragraph{Segmentation.}
Patches align exactly with the model's own ViT tokenisation: for a model with patch size $p$, the
image is divided into an $(H/p) \times (W/p)$ non-overlapping grid using a \texttt{GridSegmenter}
whose cell size is read directly from \texttt{patch\_embed.proj.kernel\_size}. This ensures each
occlusion unit corresponds to one input token and avoids any resolution mismatch between the
attribution granularity and the model's processing granularity. A pixel-level variant
(\texttt{PixelSegmenter}, percentile-based bins) is used in ablation experiments to verify that the
results are not sensitive to segmentation granularity, rankings were consistent across all models
and methods.

\paragraph{Background imputation.}
Occluded patches are filled with the per-channel ImageNet mean, applied identically under both MIF
and LIF protocols. This choice follows the recommendation in~\cite{blucher2024decoupling}: using
the dataset mean avoids introducing out-of-distribution low-level statistics while keeping the fill
value independent of the sample, which is necessary for the LIF/MIF comparison to be unconfounded.
A sample-mean imputer (replacing occluded tokens with their own per-channel average) was evaluated
as an ablation; results were qualitatively consistent.

\paragraph{Tie-breaking.}
When multiple patches share identical attribution scores (common for sparse or quantized methods),
their relative ordering is ambiguous. We break ties by adding a small uniform perturbation
$\varepsilon \sim \mathcal{U}[0,\,\delta/2]$ to each patch score, where $\delta$ is half the
smallest non-zero absolute difference between any two distinct scores in the sample. This
perturbation never exceeds $10^{-6}$ and therefore cannot alter any valid ordering; it only
resolves ambiguities among truly tied entries.

\paragraph{Scoring variant.}
We report \emph{logit \gls{srg}}: the curves record the raw output logit $z_y(\cdot)$ rather than
the softmax-normalised probability. This avoids the compression effect of the softmax near $0$ and
$1$, making the metric more sensitive to small changes in the model's relative class preference.
Prediction-probability \gls{srg} (using $\mathrm{softmax}(z)_y$ instead) was also computed and
yields consistent rankings.

\subsection{Localization Metric: Attribution Localization}
\label{sec:localization}

Faithfulness (\gls{srg}) measures whether the attribution correctly predicts what the model
responds to. It says nothing about whether those regions correspond to the semantically relevant
object in the image. We therefore complement \gls{srg} with a \emph{spatial localization} metric
that evaluates how well attribution maps agree with pixel-level ground-truth segmentation masks,
independent of model behaviour.

\paragraph{Metric definition.}
Given an attribution map $A \in \mathbb{R}^{H \times W}$ (obtained by summing over channels) and a
binary segmentation mask $\mathcal{S} \subseteq \{1,\ldots,H\} \times \{1,\ldots,W\}$, the
\emph{Attribution Localization} score is the fraction of total absolute attribution mass inside the
object mask:
\begin{equation}
  \label{eq:loc}
  \mathrm{Loc}(A, \mathcal{S})
  \;=\; \frac{\textstyle\sum_{(i,j)\in\mathcal{S}} |A_{ij}|}
  {\textstyle\sum_{i,j} |A_{ij}|}.
\end{equation}
A score of $1$ means all attribution is inside the object;
a score equal to the relative mask area is the expected value for a
spatially uniform (uninformative) attribution map.
We use the implementation from the Quantus library~\cite{hedstrom2023quantus} with
all attribution values clipped with a minimal value of $0$ (\texttt{positive\_attributions=True})
so that methods with signed outputs are treated symmetrically with
those that produce strictly non-negative maps.

\paragraph{Dataset.}
Localization is evaluated on ImageNet-S~\citep{gao2022luss}, a subset of ImageNet-1k whose
validation images come with high-quality pixel-level segmentation masks. For each model we select
50~classes uniformly at random (seed~754068) and retain the first 10~\emph{correctly classified}
samples per class, yielding 500~evaluation images in total. Restricting to correctly classified
samples ensures that the model's prediction and the segmentation mask refer to the same object: an
image that the model misclassifies could legitimately produce attributions focused on a secondary
object, which would be penalised unfairly by the localization metric. If the attribution map has a
different spatial resolution than the mask (e.g.\ due to a different patch size), it is resized to
the mask dimensions with bilinear interpolation before computing Equation~\eqref{eq:loc}.

\subsection{FunnyBirds Protocol}
\label{app:funnybirds}

\paragraph{Dataset.}
FunnyBirds~\cite{hesse2023funnybirds} is a synthetic dataset of 50 bird classes rendered at
$256\times256$ resolution, with 50{,}000 training and 5{,}000 test images. Every bird is assembled
from an inventory of 26 predefined parts grouped into five human-comprehensible categories, namely
beak, wings, feet, eyes, and tail. Because each scene is rendered rather than photographed, any
subset of parts can be removed and the bird re-rendered in distribution. Comparing the model's
output before and after such an intervention gives ground-truth part importance directly, which is
precisely what \gls{srg} and ImageNet-S localization cannot supply.

\paragraph{Setup.}
We use the official dataset, the official metric implementations, and the ViT-B/16 checkpoint
released by \citet{hesse2023funnybirds}, so our scores are directly comparable to the published
baselines. The framework requires an attribution method to expose two entry points.
\texttt{get\_part\_importance} returns one scalar per bird part, and \texttt{get\_important\_parts}
returns the set of parts judged important, swept over roughly 80 thresholds. We obtain both by
summing the attribution map over the renderer's per-part segmentation masks.

\paragraph{Metrics.}
All six reported scores lie in $[0,1]$ and higher is better. Four constitute the
\emph{completeness} dimension. The controlled synthetic data check (CSDC) evaluates agreement on
scenes constructed so that the causally relevant parts are known by design. The preservation check
(PC) and the deletion check (DC) test whether the prediction survives when only the parts marked
important are kept, and whether it changes when exactly those parts are removed. Distractibility
(Distr.) measures whether parts irrelevant to the class are correctly left unattributed.
\emph{Correctness} is measured by single deletion (SD), the Spearman rank correlation between the
attributed part importances and the change in the target output when each part is removed on its
own, which makes SD the metric that compares an explanation against intervention-derived ground
truth most directly. \emph{Contrastivity} is measured by target sensitivity (TS), which checks that
the explanation is specific to the explained class rather than shared across classes. We omit
background independence, which the framework reports separately and which probes the backbone
rather than the attribution rule.

\subsection{SAE Training and Evaluation Details}
\label{sec:sae_details}

\begin{table}[t]
  \centering
  \caption{\glsentryshort{sae} Latent \glsentryshort{srg} Benchmark. \glsentryshort{srg}: higher is better; Rank: lower is better (1 = best). Bold entries are within 1 standard error of the best.}
  \label{tab:sae_srg}
  \small
  \setlength{\tabcolsep}{4pt}
  \resizebox{\textwidth}{!}{%
    \begin{tabular}{l|ccc|ccc}
      \toprule
      \textbf{Method}         & \textbf{blocks.15} \glsentryshort{srg} & \textbf{blocks.19} \glsentryshort{srg} & \textbf{blocks.23} \glsentryshort{srg} & \textbf{blocks.15} Rank    & \textbf{blocks.19} Rank    & \textbf{blocks.23} Rank    \\
      \midrule
      \glsentryshort{ours}    & \textbf{0.449 $\pm$ 0.006}             & \textbf{0.366 $\pm$ 0.006}             & \textbf{0.227 $\pm$ 0.006}             & \textbf{1.597 $\pm$ 0.028} & \textbf{2.018 $\pm$ 0.034} & \textbf{2.364 $\pm$ 0.039} \\
      \glsentryshort{attnlrp} & 0.334 $\pm$ 0.006                      & 0.322 $\pm$ 0.006                      & 0.207 $\pm$ 0.005                      & 2.983 $\pm$ 0.042          & 2.589 $\pm$ 0.043          & 2.621 $\pm$ 0.039          \\
      \glsentryshort{cplrp}   & 0.249 $\pm$ 0.005                      & 0.238 $\pm$ 0.005                      & 0.176 $\pm$ 0.004                      & 3.972 $\pm$ 0.036          & 3.673 $\pm$ 0.040          & 3.040 $\pm$ 0.040          \\
      \glsentryshort{actmap}  & 0.279 $\pm$ 0.006                      & 0.233 $\pm$ 0.006                      & -0.031 $\pm$ 0.006                     & 3.696 $\pm$ 0.040          & 3.636 $\pm$ 0.043          & 4.294 $\pm$ 0.038          \\
      \bottomrule
    \end{tabular}%
  }
\end{table}

\paragraph{Model.}
SAEs are trained on activations of the ViT-L/14 A backbone (see \cref{tab:models}).

\paragraph{Architecture.}
Each \gls{sae} operates on the token level, processing all tokens (including the CLS token): the
batch and sequence dimensions of the activation tensor are flattened so that each token is
processed independently. The SAE maps each 1024-dimensional token vector to a 32,768-dimensional
sparse code (expansion factor 32) and back, enforcing $k=64$ active features per token via the TopK
activation function.

\paragraph{Training.}
Training uses Adam ($\text{lr} = 3\times10^{-4}$, seed $23001$) with a cosine warm-up schedule (100
warm-up steps) and gradient clipping at $1.0$, on the full ImageNet-1k training set with an
effective batch size of $512$ (32 per device $\times$ 16 gradient accumulation steps). Dead
features are periodically reanimated via an auxiliary loss term with coefficient $10^{-3}$, which
adds a penalty proportional to the fraction of inactive features, encouraging them to recover
non-zero activations. Training runs for 3 epochs {(approx.\ 6 GPU hours on a two NVIDIA GeForce RTX
    5090)}. After training, all features are active in every layer (dead ratio $= 0$) and
reconstruction quality reaches $R^2 > 0.8$ across all layers.

\paragraph{Evaluation protocol.}
We adapt the \gls{srg} protocol (\cref{sec:srg}) by replacing the classifier logit with
$s_d(\mathbf{x})$. The $16{\times}16$ occlusion grid aligns exactly with the ViT patch tokenisation
(patch size 14\,px at 224\,px input), so each occlusion step directly corresponds to replacing one
input token with the per-channel ImageNet mean. To make scores comparable across latents with
different activation scales, curves are normalised for each (layer, latent) pair by the global
maximum across all methods and all curve steps before computing the \gls{auc}. For each evaluated
layer we select 200 strongly-activating latents and up to 5 high-activating images per latent
(${\geq}90$th percentile of nonzero activations, minimum 2 images) from the ImageNet validation
set. Results are reported as mean \gls{srg} and mean rank (1 = best) aggregated over all (latent,
image) pairs per layer.

\subsection{Evaluation Protocol for Vision-Language Models}
\label{sec:vlm_eval}

In this section we provide details on the \gls{vlm} attribution evaluation. For the metric
definitions we refer to \cref{sec:srg,sec:localization}.

\paragraph{Models and rule configuration.}
We evaluate
\begin{itemize}
  \item
        \texttt{Qwen/Qwen2.5-VL-3B-Instruct},
  \item
        \texttt{Qwen/Qwen3-VL-4B-Instruct} and
  \item
        \texttt{google/gemma-3-4b-it},
\end{itemize}
all in \texttt{bfloat16} on a single GPU without quantisation.
In all three cases the $\gamma$-rule is applied to the residual merges of the \emph{vision tower
  only} and the language model is propagated with unmodified \gls{attnlrp} ($\gamma_{\text{LLM}}=0$).
No
per-model tuning of $\gamma$, of the $\epsilon$-stabiliser or of the linear rule is performed.

\paragraph{Explained scalar.}
For each image the model is queried, and the attribution target is the logit of one answer token at
its own decoding position, back-propagated to the input pixels. Relevance reaching the image is
summed over colour channels, and the model's patchification is inverted analytically to recover a
pixel-space map $A \in \mathbb{R}^{H\times W}$. Because the image embedding and the language model
are separate sub-graphs, the backward pass is performed in two stages: relevance is first
propagated from the answer logit to the image-token embeddings, and those are then used as the
incoming relevance for a second backward pass through the vision tower to the pixels.

\paragraph{Localization.}
To obtain an unambiguous, label-aligned target token, the models are evaluated in a forced-choice
protocol. Fifty ImageNet-S classes are drawn at random (seed~754068) and each is assigned a code
string that the model's tokeniser encodes as exactly one token (single capital letters first, then
two-letter combinations, verified at run time). The prompt shows the image together with the full
code list and instructs the model to emit the code and nothing else, so that the first generated
token identifies the predicted class, cf. \cref{lst:prompt}. For each class the first ten images
the model codes \emph{correctly} are retained, giving 500 evaluation images. The attribution target
is the logit of the ground-truth class code at the first answer position. As in the \gls{vit}
experiments, restricting to correctly classified samples keeps prediction and segmentation mask
referring to the same object.

\begin{lstlisting}[caption={Evaluation prompt after chat-template
  expansion. },
  label={lst:prompt}]
  <|im_start|>system
  You are a helpful assistant.<|im_end|>
  <|im_start|>user
  <|vision_start|><|image_pad|><|vision_end|>Identify the main object in the image.
  Use ONLY the code from the list below -- output the code and nothing else.

  A = isopod
  B = Scottish deerhound
  C = ant
  ...
  AX = spoonbill<|im_end|>
  <|im_start|>assistant
  \end{lstlisting}

\paragraph{Faithfulness.}
The model is shown the image and asked \emph{``What is the main object in this photo? Answer with a
  single word.''}, generates greedily, and the attribution target is the logit of the first
\emph{content} token of its own answer, i.e.\ the first token decoding to an alphabetic
non-stopword. No correctness filter is applied, because the explained quantity is the model's own
answer rather than a fixed class. Images are resized to the vision tower's native $896\times896$
resolution, giving a fixed $64\times64$ patch grid. The evaluation set is 500 ImageNet-S validation
images spanning the same class sample.

The occlusion unit is one image token, as in \cref{sec:srg}, but \glspl{vlm} do not share a common
token geometry, so the unit differs per model. All images are resized for consistency within
models. For the Qwen models, whose processor emits a variable-length sequence of merged patch rows,
occlusion is applied directly to those rows: the per-row relevance is the sum of the pixel
relevance it carries, and occluded rows are set to the per-image mean of the preprocessed pixel
values. For Gemma-3 the fixed $896\times896$ input is partitioned into a $16\times16$ grid of
$56\times56$ pixel cells and occluded cells are set to the background value. In both cases the full
model is re-evaluated at every occlusion step, and MIF and LIF curves are read at the same answer
position and token as the attribution target. We use $T=20$ occlusion steps for the Qwen models and
$T=17$ for Gemma-3.

\paragraph{Statistics.}
All tables report the mean over the evaluation set with the standard error $\hat{\sigma}/\sqrt{n}$.
Significance is assessed per model and per metric with a two-sided Wilcoxon signed-rank test on the
per-image scores of \gls{ours} against \gls{attnlrp}, paired image by image. A paired $t$-test
gives the same conclusion in every case.

\subsection{Full Benchmark Tables}
\label{app:extended_benchmarks}

\begin{table}[htbp]
  \centering
  \caption{\glsentryshort{srg} faithfulness (patch-wise occlusion, logit target, seed 339306, $n=500$ images) across models. Values denote $\text{Mean} \pm \text{Stderr}$; higher is better. ``n.a.'' marks method/backbone combinations the method is not defined for; ``--'' marks runs that are missing. Bold: best mean. Underline: second-best mean.}
  \label{tab:srg:logit:patch:339306}
  \resizebox{\textwidth}{!}{%
    \begin{tabular}{lcccccccccc}
      \toprule
      \textbf{Method}
                                  & \textbf{ViT-B/16}
                                  & \textbf{ViT-L/14}
                                  & \textbf{ViT-H/14}
                                  & \textbf{DeiT3-L/16}
                                  & \textbf{DINOv2-S}
                                  & \textbf{DINOv2-S (Reg.)}
                                  & \textbf{DINOv2-L}
                                  & \textbf{DINOv2-L (Reg.)}
                                  & \textbf{SigLIP2-L/16}
                                  & \textbf{SwinV2-L}
      \\
      \midrule

      \glsentryshort{ours}        & \textbf{3.20 $\pm$ 0.09}    & \underline{3.53 $\pm$ 0.09} & \textbf{3.31 $\pm$
                                                                                                  0.09}          & \textbf{3.38 $\pm$ 0.08} & \textbf{6.90 $\pm$ 0.15}    & \textbf{7.12 $\pm$ 0.16} &
      \textbf{4.72 $\pm$ 0.13}    & 4.15 $\pm$ 0.14             & \textbf{5.26 $\pm$ 0.15}    & \textbf{2.36 $\pm$ 0.07}                                                                               \\

      \glsentryshort{attnlrp}     & 1.51 $\pm$ 0.07             & 1.49 $\pm$ 0.07             & 1.65 $\pm$ 0.07             & 2.69 $\pm$ 0.08          &
      6.38 $\pm$ 0.15             & 6.62 $\pm$ 0.15             & 1.65 $\pm$ 0.09             & 1.89 $\pm$ 0.10             & 0.59 $\pm$ 0.10          &
      \underline{0.93 $\pm$ 0.05}                                                                                                                                                                      \\

      \glsentryshort{cplrp}       & 1.33 $\pm$ 0.07             & 1.23 $\pm$ 0.07             & 1.33 $\pm$ 0.07             & 2.97 $\pm$ 0.08          &
      4.05 $\pm$ 0.15             & 3.68 $\pm$ 0.16             & 0.36 $\pm$ 0.08             & 0.76 $\pm$ 0.07             & 0.49 $\pm$ 0.09          & 0.80
      $\pm$ 0.04                                                                                                                                                                                       \\

      \glsentryshort{clrp}        & 2.74 $\pm$ 0.09             & 2.85 $\pm$ 0.09             & 2.35 $\pm$ 0.09             & 2.55 $\pm$ 0.08          & 5.91
      $\pm$ 0.16                  & 6.19 $\pm$ 0.16             & 4.43 $\pm$ 0.13             & \underline{4.45 $\pm$ 0.13} & 2.69 $\pm$ 0.16          &
      n.a.                                                                                                                                                                                             \\

      CheferAttnRollout           & 2.74 $\pm$ 0.09             & 2.85 $\pm$ 0.09             & 2.35 $\pm$ 0.09             & 2.65 $\pm$ 0.08          &
      \underline{6.68 $\pm$ 0.15} & \underline{6.95 $\pm$ 0.16} & \underline{4.44 $\pm$ 0.13} &
      \textbf{4.62 $\pm$ 0.13}    & n.a.                        & n.a.                                                                                                                                 \\

      GradAttnRollout             & 2.46 $\pm$ 0.09             & 2.84 $\pm$ 0.09             & 2.43 $\pm$ 0.09             & 2.87 $\pm$ 0.08          & 6.25
      $\pm$ 0.14                  & 5.85 $\pm$ 0.15             & 4.05 $\pm$ 0.12             & 4.33 $\pm$ 0.13             & n.a.                     & n.a.                                          \\

      LibraFullGrad+              & 2.72 $\pm$ 0.09             & 2.98 $\pm$ 0.09             & 2.10 $\pm$ 0.09             & 3.07 $\pm$ 0.08          & 5.77 $\pm$
      0.17                        & n.a.                        & 3.18 $\pm$ 0.13             & n.a.                        & 3.56 $\pm$ 0.16          & n.a.                                          \\

      LeGrad                      & \underline{3.18 $\pm$ 0.09} & \textbf{3.58 $\pm$ 0.09}    & \underline{3.16 $\pm$ 0.09} &
      \underline{3.12 $\pm$ 0.07} & n.a.                        & n.a.                        & n.a.                        & n.a.                     & \underline{4.20 $\pm$ 0.13} & n.a.            \\

      \glsentryshort{ig}          & 1.23 $\pm$ 0.07             & 1.36 $\pm$ 0.07             & 1.19 $\pm$ 0.07             & 1.10 $\pm$ 0.06          & 3.47
      $\pm$ 0.10                  & 3.57 $\pm$ 0.10             & 1.24 $\pm$ 0.06             & 1.13 $\pm$ 0.06             & 1.38 $\pm$ 0.07          & 0.57 $\pm$
      0.04                                                                                                                                                                                             \\

      \glsentryshort{ixg}         & 0.45 $\pm$ 0.05             & 0.33 $\pm$ 0.05             & 0.27 $\pm$ 0.04             & 0.23 $\pm$ 0.04          & 1.10
      $\pm$ 0.08                  & 1.19 $\pm$ 0.08             & 0.45 $\pm$ 0.05             & 0.41 $\pm$ 0.05             & 0.27 $\pm$ 0.06          & 0.43 $\pm$
      0.03                                                                                                                                                                                             \\

      Random                      & -0.11 $\pm$ 0.03            & -0.03 $\pm$ 0.03            & -0.04 $\pm$ 0.03            & -0.05 $\pm$ 0.03         & 0.01 $\pm$
      0.06                        & -0.04 $\pm$ 0.05            & -0.10 $\pm$ 0.03            & -0.09 $\pm$ 0.04            & -0.04 $\pm$ 0.04         & 0.02 $\pm$ 0.02
      \\ \bottomrule
    \end{tabular}%
  }
\end{table}

\begin{table}[htbp]
  \centering
  \caption{Localization (relevance mass on the target-class \gls{imagenets} mask, $n=500$ images) across models. Values denote $\text{Mean} \pm \text{Stderr}$; higher is better. ``n.a.'' marks method/backbone combinations the method is not defined for; ``--'' marks runs that are missing. Bold: best mean. Underline: second-best mean.}
  \label{tab:localization:v2targetmask}
  \resizebox{\textwidth}{!}{%
    \begin{tabular}{lcccccccccc}
      \toprule
      \textbf{Method}
                                  & \textbf{ViT-B/16}
                                  & \textbf{ViT-L/14}
                                  & \textbf{ViT-H/14}
                                  & \textbf{DeiT3-L/16}
                                  & \textbf{DINOv2-S}
                                  & \textbf{DINOv2-S (Reg.)}
                                  & \textbf{DINOv2-L}
                                  & \textbf{DINOv2-L (Reg.)}
                                  & \textbf{SigLIP2-L/16}
                                  & \textbf{SwinV2-L}
      \\
      \midrule

      \glsentryshort{ours}        & \textbf{0.62 $\pm$ 0.01}    & \textbf{0.61 $\pm$ 0.01}    & \textbf{0.61 $\pm$
                                                                                                  0.01}          & 0.55 $\pm$ 0.01             & \underline{0.63 $\pm$ 0.01} & \underline{0.60 $\pm$ 0.01} & \textbf{0.59
                                                                                                                                                                                                               $\pm$ 0.01}   & \underline{0.59 $\pm$ 0.01} & \textbf{0.51 $\pm$ 0.01} & \textbf{0.64 $\pm$ 0.01} \\

      \glsentryshort{attnlrp}     & 0.32 $\pm$ 0.01             & 0.39 $\pm$ 0.01             & 0.41 $\pm$ 0.01             & 0.44 $\pm$ 0.01             &
      0.59 $\pm$ 0.01             & 0.59 $\pm$ 0.01             & 0.47 $\pm$ 0.01             & 0.56 $\pm$ 0.01             & 0.32 $\pm$ 0.01             & 0.34
      $\pm$ 0.01                                                                                                                                                                                                                                                                                                 \\

      \glsentryshort{cplrp}       & 0.36 $\pm$ 0.01             & 0.46 $\pm$ 0.01             & 0.41 $\pm$ 0.01             & 0.54 $\pm$ 0.01             &
      0.51 $\pm$ 0.01             & 0.48 $\pm$ 0.01             & 0.44 $\pm$ 0.01             & 0.45 $\pm$ 0.01             & 0.32 $\pm$ 0.01             & 0.32
      $\pm$ 0.01                                                                                                                                                                                                                                                                                                 \\

      \glsentryshort{clrp}        & \textbf{0.62 $\pm$ 0.01}    & \underline{0.51 $\pm$ 0.01} & 0.46 $\pm$ 0.01             &
      \underline{0.69 $\pm$ 0.01} & \textbf{0.67 $\pm$ 0.01}    & \textbf{0.67 $\pm$ 0.01}    & \underline{0.58
                                                                                                  $\pm$ 0.01}             & \textbf{0.63 $\pm$ 0.01}    & 0.29 $\pm$ 0.01             & n.a.                                                                                                                     \\

      CheferAttnRollout           & 0.50 $\pm$ 0.01             & 0.46 $\pm$ 0.01             & 0.44 $\pm$ 0.01             & 0.61 $\pm$ 0.01             & 0.60
      $\pm$ 0.01                  & 0.57 $\pm$ 0.01             & 0.51 $\pm$ 0.01             & 0.55 $\pm$ 0.01             & n.a.                        & n.a.                                                                                                                                                 \\

      GradAttnRollout             & 0.50 $\pm$ 0.01             & 0.49 $\pm$ 0.01             & \underline{0.47 $\pm$ 0.01} & \textbf{0.70
                                                                                                                                $\pm$ 0.01}               & 0.59 $\pm$ 0.01             & 0.55 $\pm$ 0.01             & 0.52 $\pm$ 0.01 & 0.58 $\pm$ 0.01             & n.a.                     & n.a.
      \\

      LibraFullGrad+              & \underline{0.54 $\pm$ 0.01} & 0.49 $\pm$ 0.01             & 0.43 $\pm$ 0.01             & 0.54 $\pm$ 0.01
                                  & 0.56 $\pm$ 0.01             & n.a.                        & 0.48 $\pm$ 0.01             & n.a.                        & \underline{0.37 $\pm$ 0.01} & n.a.                                                                                                                   \\

      LeGrad                      & 0.53 $\pm$ 0.01             & 0.47 $\pm$ 0.01             & 0.46 $\pm$ 0.01             & 0.51 $\pm$ 0.01             & n.a.                        & n.a.                        & n.a.
                                  & n.a.                        & 0.33 $\pm$ 0.01             & n.a.                                                                                                                                                                                                             \\

      \glsentryshort{ig}          & 0.45 $\pm$ 0.01             & 0.38 $\pm$ 0.01             & 0.31 $\pm$ 0.01             & 0.30 $\pm$ 0.01             & 0.49
      $\pm$ 0.01                  & 0.48 $\pm$ 0.01             & 0.43 $\pm$ 0.01             & 0.45 $\pm$ 0.01             & 0.27 $\pm$ 0.01             & 0.35 $\pm$
      0.01                                                                                                                                                                                                                                                                                                       \\

      \glsentryshort{ixg}         & 0.40 $\pm$ 0.01             & 0.26 $\pm$ 0.01             & 0.22 $\pm$ 0.01             & 0.25 $\pm$ 0.01             & 0.46
      $\pm$ 0.01                  & 0.46 $\pm$ 0.01             & 0.42 $\pm$ 0.01             & 0.43 $\pm$ 0.01             & 0.22 $\pm$ 0.01             & 0.31 $\pm$
      0.01                                                                                                                                                                                                                                                                                                       \\

      Random                      & 0.37 $\pm$ 0.01             & 0.34 $\pm$ 0.01             & 0.34 $\pm$ 0.01             & 0.34 $\pm$ 0.01             & 0.42 $\pm$ 0.01             &
      0.41 $\pm$ 0.01             & 0.40 $\pm$ 0.01             & 0.40 $\pm$ 0.01             & 0.29 $\pm$ 0.01             & \underline{0.39 $\pm$ 0.01}
      \\ \bottomrule
    \end{tabular}%
  }
\end{table}

\subsection{\texorpdfstring{{Sensitivity to the $\gamma$ Parameter}}{Sensitivity to the gamma Parameter}}
\label{sec:gamma}

{We sweep $\gamma \in \{0, 0.25, 0.5, 0.75, 1, 1.25, 1.5, 2, 3, 5, 10\}$ on six models with
  both \gls{srg} and ImageNet-S localization. Every $\gamma > 0$ outperforms the \gls{attnlrp}
  baseline, recovered at $\gamma = 0$, on all models and both metrics. Performance rises sharply to a
  universal knee near $\gamma \approx 0.25$ and then plateaus. As summarized in
  \cref{tab:gamma_sweep}, the untuned default $\gamma = 1$ attains 98--100\% of the best-$\gamma$
  \gls{srg} for 5/7 models (80--90\% for the remaining two) and 92--100\% of the best-$\gamma$
  localization for all models. Increasing $\gamma$ beyond 1 changes metrics only marginally in either
  direction, so no careful tuning is required. This is theoretically expected. As stated in
  \cref{methods:boundness}, $\gamma$ monotonically tightens the amplification bound
  $(1+2/\gamma)^{2L}$ while interpolating between standard \gls{lrp} and a non-amplifying,
  sign-consistent redistribution.}

{\begin{table}[t]
    \centering
    \caption{{$\gamma$ sensitivity summary, reported as \glsentryshort{srg} (logit) / ImageNet-S
        localization. Best denotes the best value over the sweep with its $\gamma$ in parentheses.
        $\gamma = 0$ recovers \glsentryshort{attnlrp}. $n=100$}}
    \label{tab:gamma_sweep}

    \begin{tabular}{lcccc}
      \toprule
      Model     & $\gamma=0$   & $\gamma=1$   & Best ($\gamma$)           & $\gamma$=1/Best \\
      \midrule
      ViT-B/16  & 2.29 / 0.342 & 3.74 / 0.589 & 3.77 (2) / 0.596 (5)      & 98\% / 97\%     \\
      ViT-L/14  & 1.49 / 0.393 & 3.53 / 0.611 & 3.58 (3) / 0.612 (1.25)   & 98\% / 100\%    \\
      ViT-H/14  & 1.65 / 0.407 & 3.31 / 0.609 & 3.32 (1.25) / 0.611 (0.5) & 99\% / 99\%     \\
      SigLIP2-L & 0.61 / 0.321 & 5.47 / 0.506 & 5.57 (5) / 0.510 (2)      & 98\% / 98\%     \\
      DeiT3-M   & 1.72 / 0.510 & 2.62 / 0.595 & 2.72 (10) / 0.595 (0.75)  & 90\% / 100\%    \\
      SwinV2-L  & 0.93 / 0.344 & 2.36 / 0.638 & 2.71 (10) / 0.663 (10)    & 80\% / 92\%     \\
      \bottomrule
    \end{tabular}
  \end{table}}

\subsection{{Registers Stabilize the Residual Stream}}
\label{sec:registers}

{Beyond improving attribution, our amplification measure $\mathcal{A}$ serves as an
  architecture-level diagnostic. Register tokens were introduced to suppress attention
  artifacts~\cite{Darcet2023VisionTN}. Using $\mathcal{A}$, we find that they also mitigate
  cancellation in the residual stream, an architectural effect previous attribution methods could not
  observe, as none defines a residual amplification measure. Across four matched DINOv2 pairs
  ($n=500$), registers consistently reduce residual amplification, and accordingly the models that
  need \gls{ours} most are those without registers (\cref{tab:registers}). Hence, registers stabilize the residual stream itself, and $\mathcal{A}$ predicts
  where attribution degrades. This prediction extends to \glspl{vlm}, which lack register-style
  mitigation, show the strongest residual cancellation we measured, and, as predicted, the largest
  gains from \gls{ours}, as we quantify next.}

\begin{table}[b]
  \centering
  \caption{{Register tokens reduce residual amplification $\mathcal{A}$ and, accordingly, the
      localization gain of \glsentryshort{ours} over \glsentryshort{attnlrp}, measured on four matched DINOv2 pairs
      ($n=500$). Full benchmark results in \cref{app:extended_benchmarks}.}}
  \label{tab:registers}

  \begin{tabular}{lcc}
    \toprule
    DINOv2 & $\mathcal{A}$ (w/o $\to$ w/ registers) & \glsentryshort{ours} localization gain (w/o $\to$ w/ registers) \\
    \midrule
    S      & $1.58 \to 1.45$                        & $+0.037 \to +0.016$                                             \\
    B      & $1.49 \to 1.39$                        & $+0.133 \to +0.034$                                             \\
    L      & $1.32 \to 1.28$                        & $+0.132 \to +0.040$                                             \\
    G      & $1.12 \to 1.09$                        & $+0.026 \to +0.000$                                             \\
    \bottomrule
  \end{tabular}
\end{table}

\subsection{Details on the Causal Analyses}
\label{app:causal_details}

\paragraph{Intervention protocol.}
For each residual merge, channels are ranked by their cancellation score $\mathcal{C}$ computed per
channel over a held-out calibration set. The residual $\gamma$-rule is then applied only to the
selected channel subset (top-ranked, bottom-ranked, or random), while all remaining channels use
the standard propagation rule. Relevance covered denotes the fraction of exploding relevance mass
affected by the intervention. \gls{srg} is evaluated with the protocol of \cref{sec:srg}.

\paragraph{Replication on DINOv2-B.}
The intervention pattern reported for ViT-B/16 in \cref{tab:interventions}a replicates on DINOv2-B
in \cref{tab:interventions_dino}b. Correcting only the top 2\% most cancellation-prone channels
improves \gls{srg} by +9.9\%, while treating 75\% of low-cancellation channels yields only +8.1\%.

\begin{table}[h]
  \centering
  \caption{{Channel-wise intervention study. The residual $\gamma$-rule is applied only to the
      selected channels. Relevance covered denotes the fraction of exploding relevance mass affected
      by the intervention. \textbf{(a)} ViT-B/16. \textbf{(b)} DINOv2-B, showing the same pattern.}}
  \begin{subtable}[t]{0.49\textwidth}
    \centering
    \label{tab:interventions}
    \textbf{(a)}\\[2pt]
    \resizebox{\textwidth}{!}{%
      \begin{tabular}{lccc}
        \toprule
        ViT-B/16 intervention       & $\gamma$-treated & Relevance covered & \glsentryshort{srg} $\uparrow$ \\
        \midrule
        \glsentryshort{attnlrp}     & 0\%              & 0\%               & 0.127                          \\
        Top cancellation            & 0.5\% (92)       & 21\%              & 0.137 (+7.9\%)                 \\
        Top cancellation            & 20\% (3{,}686)   & 75\%              & 0.168 (+32\%)                  \\
        Top cancellation            & 75\% (13{,}824)  & 97\%              & 0.313 (+146\%)                 \\
        Random                      & 75\% (13{,}824)  & 76\%              & 0.252 (+98\%)                  \\
        Bottom cancellation         & 75\% (13{,}824)  & 21\%              & 0.120 ($-5.5$\%)               \\
        \glsentryshort{ours} (ours) & 100\%            & 100\%             & 0.303 (+138\%)                 \\
        \bottomrule
      \end{tabular}
    }
  \end{subtable}
  \hfill
  \begin{subtable}[t]{0.49\textwidth}
    \centering
    \label{tab:interventions_dino}
    \textbf{(b)}\\[2pt]
    \resizebox{\textwidth}{!}{%
      \begin{tabular}{lccc}
        \toprule
        DINOv2-B intervention & $\gamma$-treated & Relevance covered & \gls{srg} $\uparrow$ \\
        \midrule
        Attn\gls{lrp}         & 0\%              & 0\%               & 0.334                \\
        Top cancellation      & 2\% (368)        & 23\%              & 0.367 (+9.9\%)       \\
        Top cancellation      & 50\% (9{,}216)   & 84\%              & 0.403 (+20.7\%)      \\
        Top cancellation      & 75\% (13{,}824)  & 96\%              & 0.430 (+28.7\%)      \\
        Random                & 75\% (13{,}824)  & 77\%              & 0.415 (+24.3\%)      \\
        Bottom cancellation   & 75\% (13{,}824)  & 36\%              & 0.361 (+8.1\%)       \\
        \gls{ours} (ours)     & 100\%            & 100\%             & 0.425 (+27.2\%)      \\
        \bottomrule
      \end{tabular}
    }
  \end{subtable}
\end{table}

\paragraph{Component ablations.}
\Cref{tab:component_ablations} reports \gls{srg} when disabling the Attn\gls{lrp} attention or
LayerNorm rules. Both components are integral to the framework, but \gls{ours} remains
substantially more faithful than Attn\gls{lrp} under both ablations. The residual $\gamma$-rule
therefore provides a robust gain beyond the existing rules.

\begin{table}[h]
  \centering
  \caption{{LRP-component ablations, reported as \gls{srg} (logit). Percentages denote the
      relative change with respect to the full rule set.}}
  \label{tab:component_ablations}

  \begin{tabular}{lccc}
    \toprule
    \gls{srg} (logit)       & All rules & w/o Attn            & w/o LN         \\
    \midrule
    SigLIP2: Attn\gls{lrp}  & 1.01      & $-0.004$ ($-100$\%) & 0.29 ($-71$\%) \\
    SigLIP2: \gls{ours}     & 4.75      & 3.27 ($-31$\%)      & 0.78 ($-84$\%) \\
    DINOv2-S: Attn\gls{lrp} & 6.32      & 0.61 ($-90$\%)      & 1.58 ($-75$\%) \\
    DINOv2-S: \gls{ours}    & 6.83      & 3.79 ($-45$\%)      & 1.74 ($-74$\%) \\
    \bottomrule
  \end{tabular}
\end{table}

\subsection{\texorpdfstring{Qualitative \gls{ours} Attribution Examples}{Qualitative ResiLRP Attribution Examples}} \label{app:qualitative_examples}

\begin{figure}[t]
  \centering
  \includegraphics[width=0.95\linewidth]{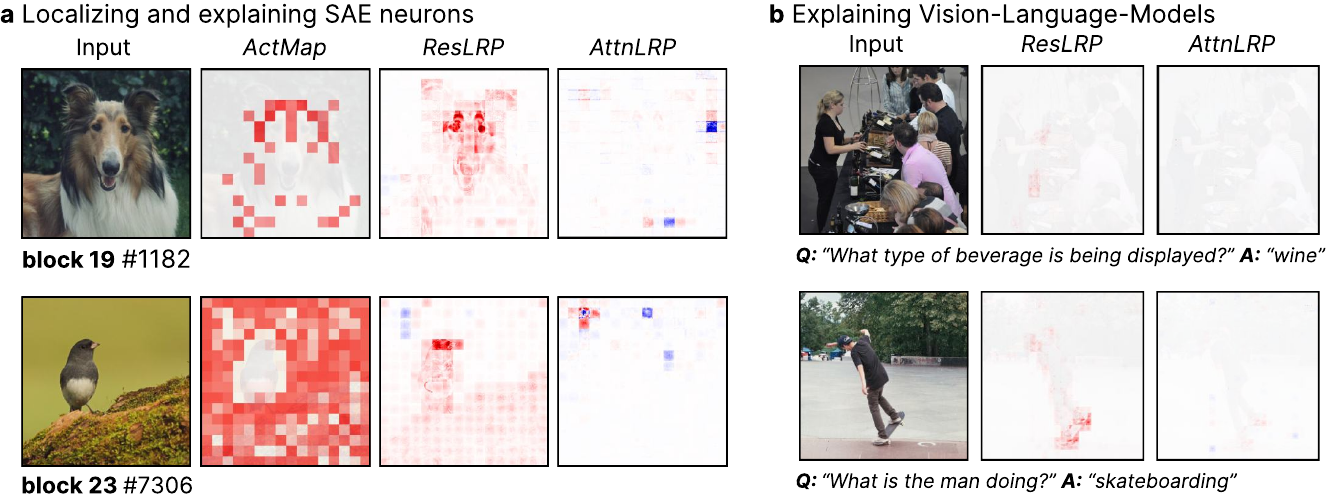}
  \caption{Additional qualitative examples of \gls{ours} attributions.
    \textbf{a)} \Gls{sae} activation maps (\mbox{ActMap}) do not always spatially align with
    the encoded semantic concepts (``dog eyes'' and ``bird''), whereas \gls{ours} highlights
    input features faithful to the \gls{sae}'s learned latent encoder (cf.\ \cref{tab:sae_srg}).
    \textbf{b)} \Gls{ours} enables pixel-level grounding of free-form text generation in large
    vision-language models (here Qwen2.5-VL~\cite{bai2025qwen25vltechnicalreport}).}
  \label{fig:app:qualitative_qwen25}
\end{figure}

\begin{figure}[t]
  \centering
  \includegraphics[width=0.99\linewidth]{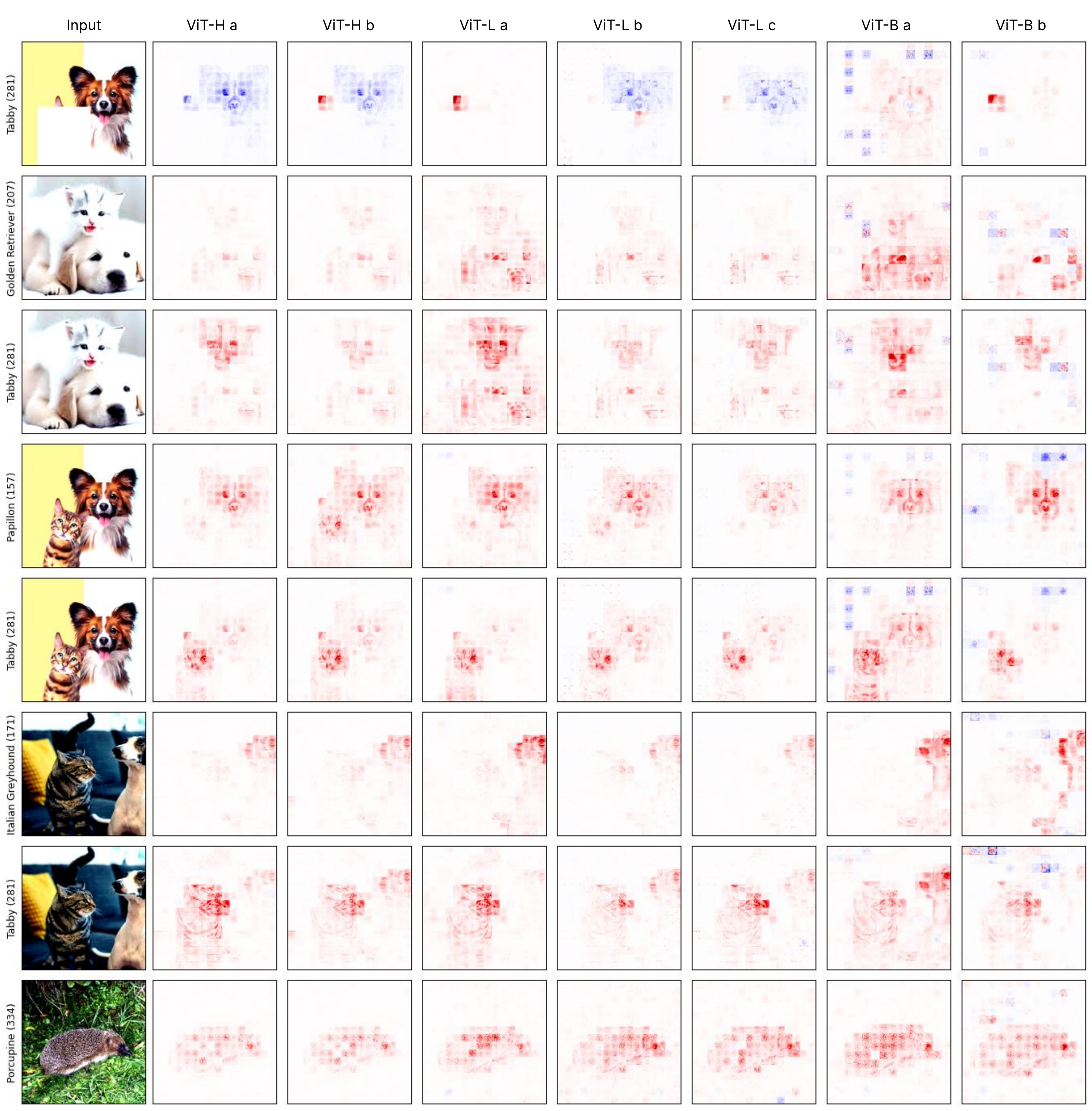}
  \caption{Qualitative examples of \gls{ours} attribution maps for different \glspl{vit} models. Explanation target is displayed on the left.}
  \label{fig:app:qualitative_1}
\end{figure}

\begin{figure}[t]
  \centering
  \includegraphics[width=0.99\linewidth]{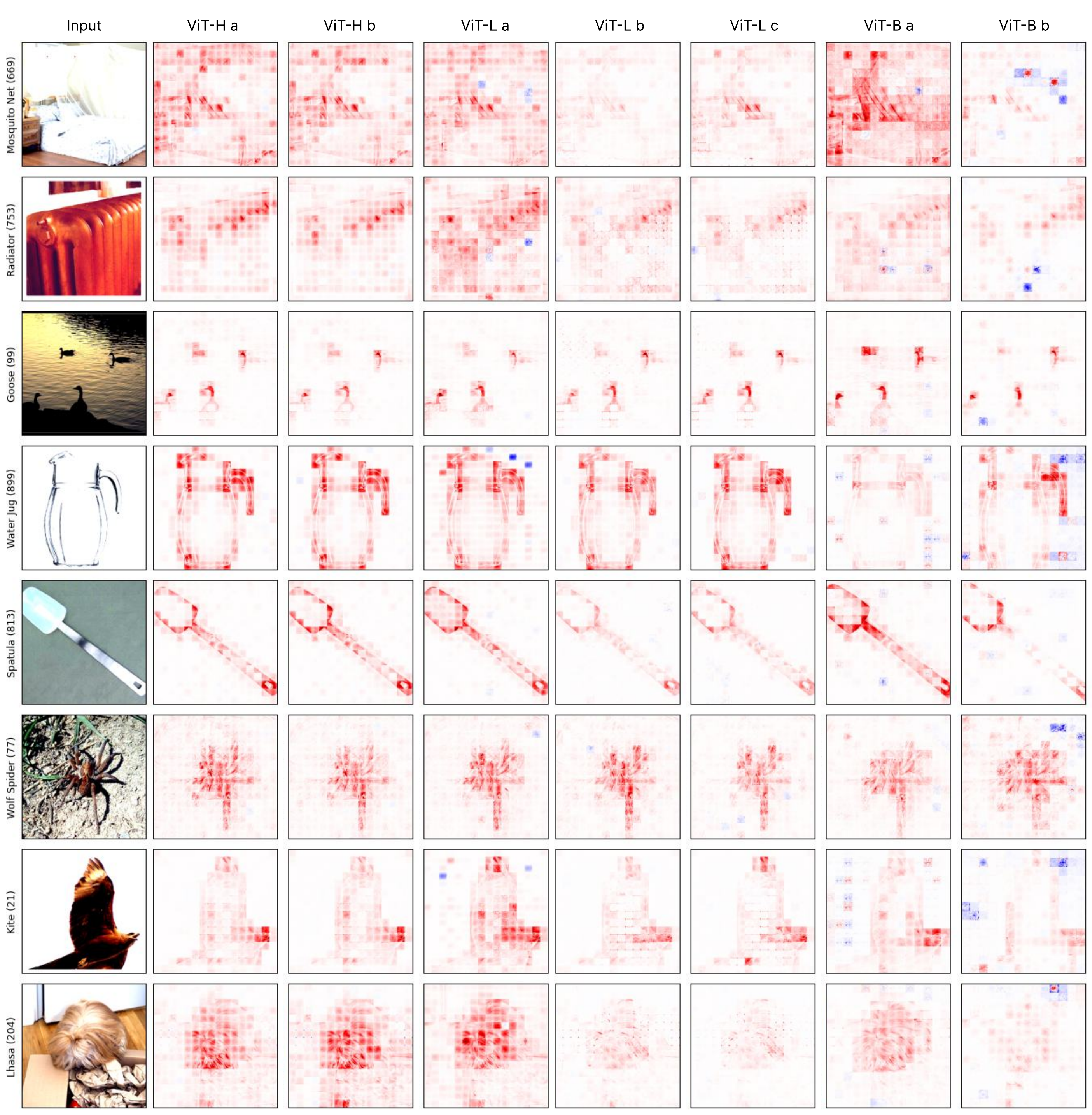}
  \caption{Qualitative examples of \gls{ours} attribution maps for different \glspl{vit} models. Explanation target is displayed on the left.}
  \label{fig:app:qualitative_2}
\end{figure}

Qualitative examples of \gls{ours} (named \gls{attnlrp}-Gamma=1) attribution maps for different
\glspl{vit} models are presented in \cref{fig:app:qualitative_1,fig:app:qualitative_2}. These
examples correspond to \cref{sec:quantitative} of the main manuscript.
\Cref{fig:app:qualitative_qwen25} additionally shows \gls{sae} feature localization and token-level
grounding for Qwen2.5-VL, complementing \cref{fig:qualitative}.

\newpage \newpage

\end{document}